\documentclass{article}
\usepackage{ijcai24}

\usepackage{times}
\usepackage{soul}
\usepackage{url}
\usepackage[hidelinks]{hyperref}
\usepackage[utf8]{inputenc}
\usepackage[small]{caption}
\usepackage{graphicx}
\usepackage{amsmath}
\usepackage{amsthm}
\usepackage{booktabs}
\usepackage{algorithm}
\usepackage{algorithmic}
\usepackage[switch]{lineno}

\usepackage{dsfont}
\usepackage{pdfpages}
\usepackage{booktabs} 
\usepackage{siunitx} 
\usepackage{multirow} 
\usepackage{amssymb} 
\usepackage{caption} 
\usepackage{xspace}
\usepackage[]{subcaption} 
\usepackage{tabularx} 
\newcolumntype{R}{>{\raggedleft\arraybackslash}X}
\usepackage{bm} 
\usepackage{makecell}
\usepackage[export]{adjustbox}
\usepackage{newfloat}

\newcommand{\ie}{\emph{i.e.}}

\title{Revealing Hierarchical Structure of Leaf Venations in Plant Science via Label-Efficient Segmentation: Dataset and Method}

\author{
    Author Name
    \affiliations
    Affiliation
    \emails
    email@example.com
}

\iftrue
\author{
    Weizhen Liu$^1$ \and
    Ao Li$^1$ \and
    Ze Wu$^1$ \and
    Yue Li$^1$ \and
    Baobin Ge$^1$ \and
    Guangyu Lan$^1$ \and
    Shilin Chen$^1$ \and \\
    Minghe Li$^2$ \and
    Yunfei Liu$^1$ \and
    Xiaohui Yuan$^1$\thanks{Corresponding authors.} \and
    Nanqing Dong$^{2*}$
    \\
    \affiliations
    $^1$Wuhan University of Technology, Wuhan, China\\
    $^2$Shanghai Artificial Intelligence Laboratory, Shanghai, China
    \\
    \emails
    \{liuweizhen, liao2022, arlowu313, gebaobin555, languangyu, chenshilin0125, yuanxiaohui\}@whut.edu.cn,
    \{liminghe, dongnanqing\}@pjlab.org.cn
}
\fi
\begin{document}

\maketitle
\begin{abstract}
Hierarchical leaf vein segmentation is a crucial but under-explored task in agricultural sciences, where analysis of the hierarchical structure of plant leaf venation can contribute to plant breeding. While current segmentation techniques rely on data-driven models, there is no publicly available dataset specifically designed for hierarchical leaf vein segmentation. 
To address this gap, we introduce the HierArchical Leaf Vein Segmentation (HALVS) dataset, the first public hierarchical leaf vein segmentation dataset. HALVS comprises 5,057 real-scanned high-resolution leaf images collected from three plant species: soybean, sweet cherry, and London planetree. It also includes human-annotated ground truth for three orders of leaf veins, with a total labeling effort of 83.8 person-days. Based on HALVS, we further develop a label-efficient learning paradigm that leverages partial label information, \ie~missing annotations for tertiary veins. Empirical studies are performed on HALVS, revealing new observations, challenges, and research directions on leaf vein segmentation.
\end{abstract}

\section{Introduction}
Analyzing the detailed structure at different hierarchical levels of leaf venation is a fundamental step for botanists, crop breeders, and ecologists to understand the impact of these intricate structures on important physiological functions of leaves, such as photosynthesis, transpiration, respiration, and transportation. 
This understanding can facilitate the breeding of plant species with high yield, quality, or economic value~\cite{sack2013crop}, thus not only targeting the United Nations’ Sustainable Development Goals of No Poverty and Zero Hunger (SDG1 \& SDG2)~\cite{UN2023sdg} but also fulfilling the Leave No One Behind Principle (LNOB)~\cite{UN2023lnob}.
\begin{figure}[t]
    \centering
    \captionsetup{font=footnotesize}
  \begin{subfigure}[t]{0.23\columnwidth}
    \centering
    \includegraphics[width=\columnwidth]{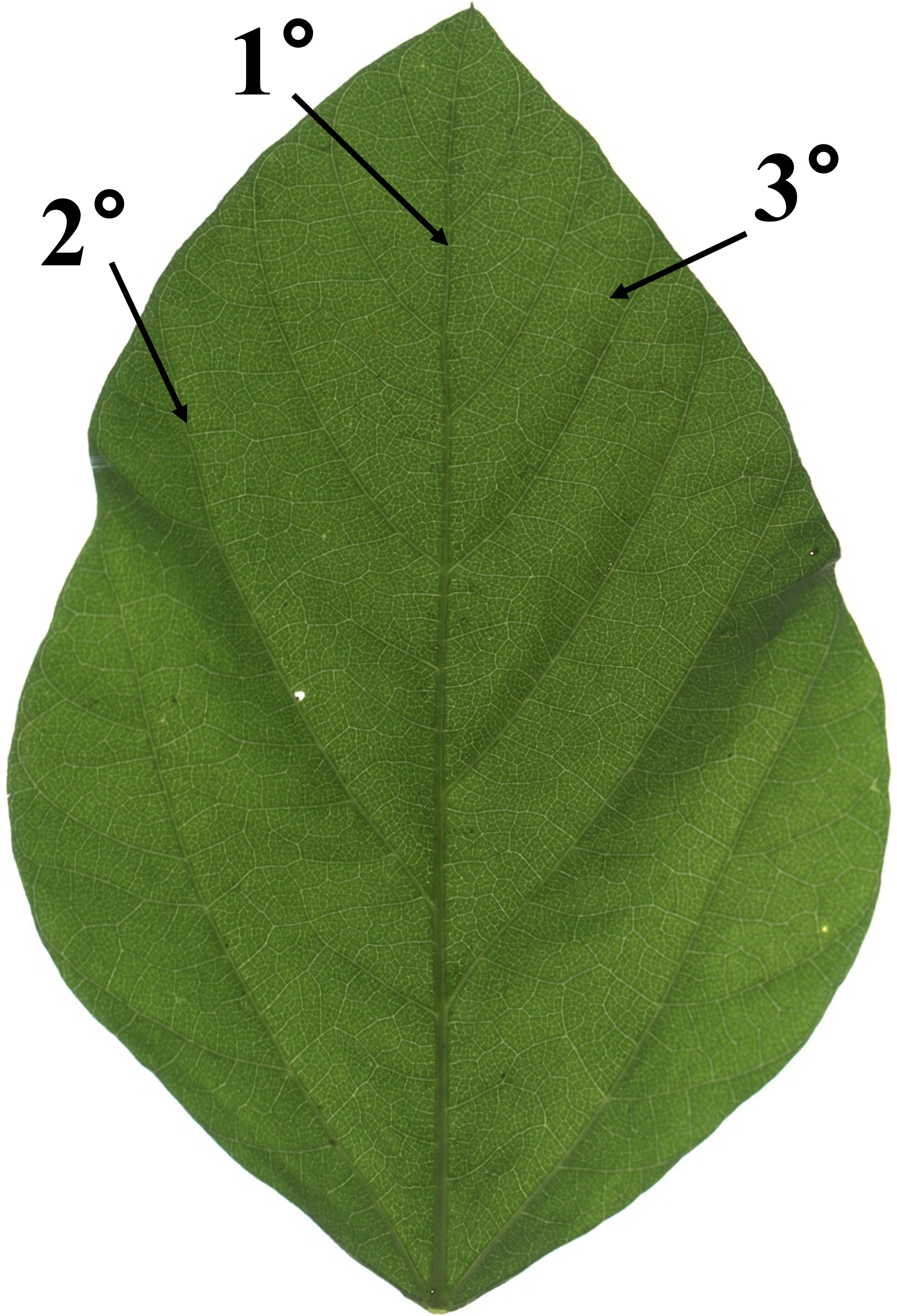}
    \captionsetup{font=footnotesize,width=1.1\textwidth}
    \caption{\centering {\small Soybean}}
    \label{fig:annotation_description_1}
  \end{subfigure}%
  \hspace{10pt}
  \begin{subfigure}[t]{0.23\columnwidth}
    \centering
    \includegraphics[width=\columnwidth]{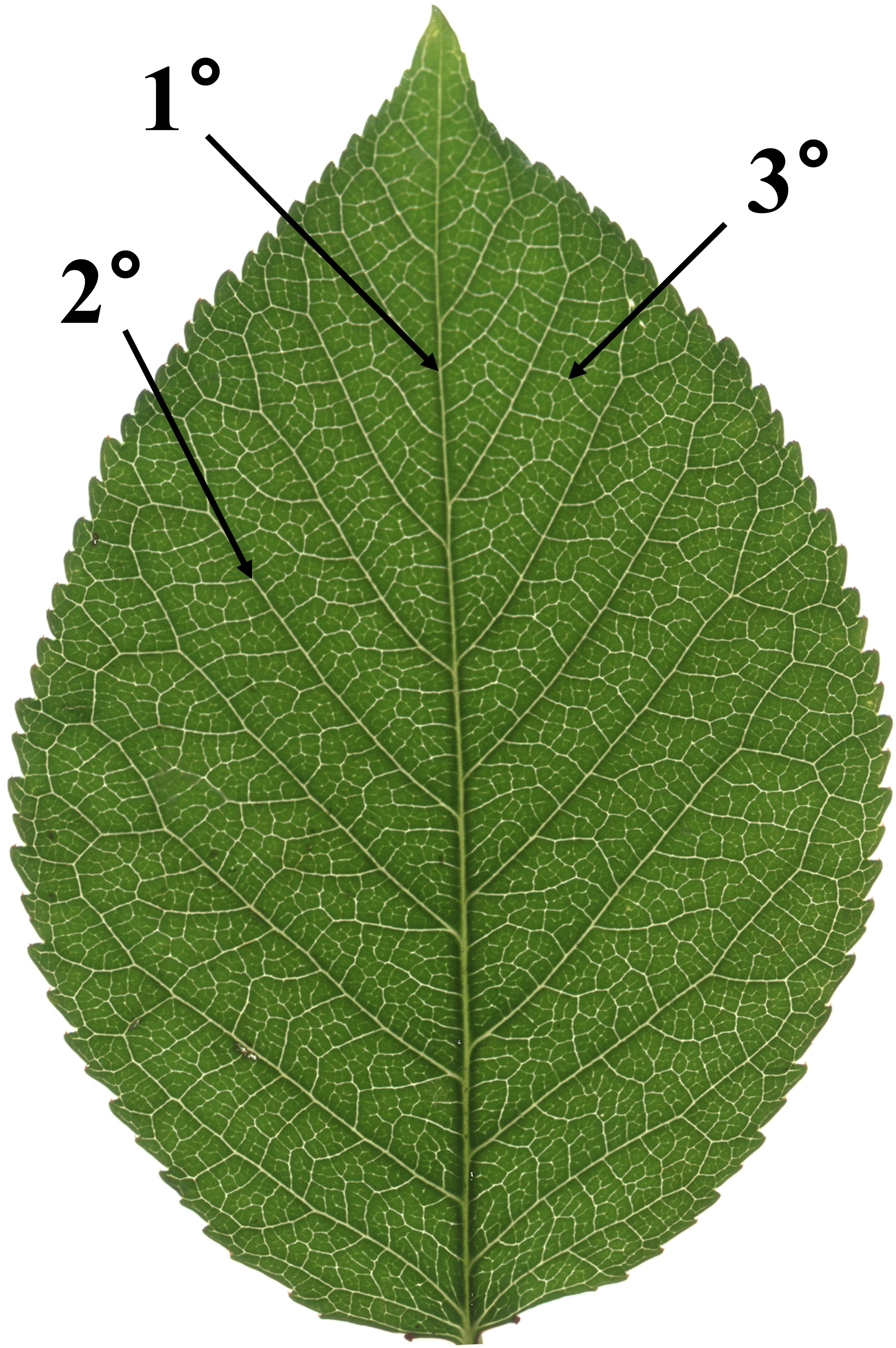}
    \captionsetup{font=footnotesize,width=1.15\textwidth}
    \caption{\centering {\small Sweet cherry}}
    \label{fig:annotation_description_2}
  \end{subfigure}%
    \hspace{10pt}
  \begin{subfigure}[t]{0.33\columnwidth}
    \centering
    \includegraphics[width=\columnwidth]{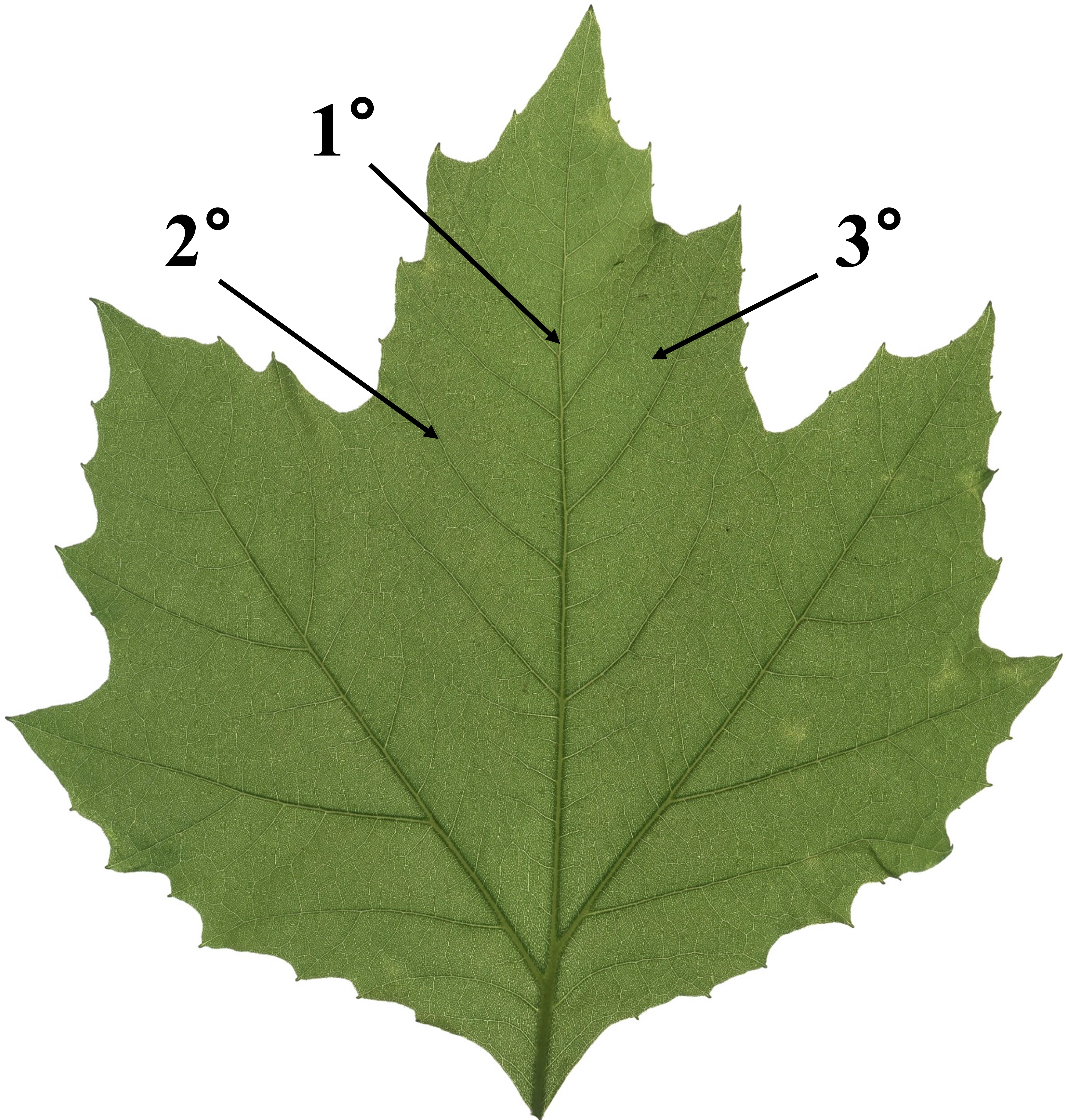}
    \captionsetup{font=footnotesize,width=1.1\textwidth}
    \caption{\centering {\small London planetree}}
    \label{fig:annotation_description_3}
  \end{subfigure}
  \caption{Illustration of three orders of veins (1\textdegree, 2\textdegree, and 3\textdegree) for three species of interest in HALVS.
  1\textdegree\ veins, typically the thickest veins in the leaf, appear as the mid-vein of the leaf. 2\textdegree\ veins are the veins of the next smaller size that branch off from the 1\textdegree\ veins. 3\textdegree\ veins are the subsequent finer branches that stem from the 2\textdegree\ veins. Best viewed with digital zoom.}
  \label{fig:annotation_description}
\end{figure}
To achieve this task, a non-negligible process is to segment the leaf venation. With the development of imaging technology and computer vision techniques, researchers have started to utilize image analysis methods to analyze leaf venation structures. But most studies focus on segmenting the entire leaf venation network without differentiating the order of veins~\cite{price2011leaf,dhondt2012quantitative,buhler2015phenovein,salima2015leaf,grinblat2016deep,lasser2017net,xu2021automated,li2022leaf,liu2022plant,iwamasa2023network}. The task of interest, hierarchical vein segmentation from leaf images, involves not only isolating and identifying the venation network from other leaf components but also classifying each order of veins into distinct semantic categories. It is a challenging computer vision task due to the complexity of vein structures and the high similarity in color and texture among veins of different orders~\cite{liu2022plant}. The first three orders are the primary (1\textdegree) vein or veins emerging from the leaf base to the apex, the smaller secondary (2\textdegree) veins branching at intervals from the 1\textdegree\ veins towards the leaf margin, and even finer tertiary (3\textdegree) veins~\cite{ellis2009manual}, as shown in Fig.~\ref{fig:annotation_description}. Until now, only two studies have attempted hierarchical vein segmentation ~\cite{gan2019automatic,jin2020automatic}, which used directional morphological filtering and region-growing-based methods, respectively. But these traditional digital image processing techniques cannot extract rich semantic information from the highly dense leaf venation networks and have not yet addressed the semantic segmentation of 3\textdegree\ veins.

Deep learning seems to be a promising technique for hierarchical leaf vein segmentation. However, as a data-driven method, the biggest obstacle is the lack of labeled training data. The existing public datasets have limitations in leaf image acquisition methods and vein labeling. For data collection, chemical cleaning and vein staining are commonly used practices for treating leaves before imaging to increase the color contrast between leaf veins and lamina~\cite{perez2016corrigendum,xu2021automated,iwamasa2023network}. But it is quite time-consuming and can easily damage the leaf blade. X-ray imaging makes it possible without chemical treatment but requires expensive specialized facilities~\cite{schneider2018improved,zhu2020fast}. More importantly, the existing public datasets annotate all the veins in a leaf as one semantic class \textbf{without }recognizing vein orders, which cannot serve as the ground truth for deep learning-based hierarchical leaf vein segmentation. 

To address the aforementioned limitations, in this work, we construct and release a new dataset for \textbf{H}ier\textbf{A}rchical \textbf{L}eaf \textbf{V}ein \textbf{S}egmentation (HALVS). To acquire \textbf{high-resolution} and \textbf{high-contrast} leaf venation images from the raw leaf samples, we adopt a flat-bed scanner in transmission scanning mode, which can provide backlight illumination throughout the scanning process. To the best of our knowledge, HALVS is the first public dataset that leverages this efficient and cost-effective imaging system. The HALVS comprises 5,057 leaf images from three representative species: soybean, sweet cherry, and London planetree (see Fig.~\ref{fig:annotation_description}). Each species was selected for its unique significance in diverse contexts. Soybean is widely acknowledged as a fundamental global provider of protein and oil~\cite{vollmann2016soybean} with both ecological and economic values worldwide. Sweet cherry represents a high-class fruit with significant economic value, and London planetree was selected for its relevance in urban landscaping, complementing the dataset by providing diversity in leaf venation patterns. Furthermore, we have also provided detailed pixel-level annotations for the first three orders of veins of these plants, considering that their vein traits are of significance for biological and ecological research.  It is worth noting that hierarchical annotated leaf veins are extremely time-consuming, as shown in Tab.~\ref{table:dataset_annotation_time}. The dataset contribution will be described in detail in Sec. HALVS Dataset.

\begin{table}[t]
\centering
\setlength{\tabcolsep}{1pt}
\captionsetup{justification=justified, labelfont=small, textfont=small}
{\small
\begin{tabularx}{0.47\textwidth}{@{\extracolsep{\fill}}>{\scriptsize}l>{\small}c>{\small}c>{\small}c>{\small}c@{\extracolsep{\fill}}}
\toprule
\small Species & 1\textdegree\ veins & 2\textdegree\ veins & 3\textdegree\ veins & Total Time \\
\midrule
Soybean & 14.5 ± 5.2 & \phantom{0}29.9 ± 6.4 & \phantom{0}80.8 ± 15.2 & 125.5 ± 14.2 \\
Sweet cherry & \phantom{0}5.5 ± 1.1 & \phantom{0}39.1 ± 3.7 & 164.3 ± 10.0 & 208.9 ± 10.6 \\
London planetree & 21.6 ± 2.6 & 132.6 ± 8.4 & 315.7 ± 14.5 & 469.8 ± 19.9 \\
\bottomrule
\end{tabularx}
}
\caption{Statistics of annotation time for each order of veins in each leaf image across three species. Times are measured in minutes and are presented in the format of mean ± standard deviation.}
\label{table:dataset_annotation_time}
\end{table}

This paper also makes a methodological advancement in label-efficient learning. Given the label scarcity challenge, fully supervised deep learning methods are not optimal for hierarchical leaf vein segmentation tasks. Instead, semi-supervised learning (SSL)~\cite{chapelle2009semi,papandreou2015weakly}, which learns a model with only a few labeled data and numerous unlabeled ones, looks like a feasible solution. We notice that annotations of 3\textdegree\ veins take much longer than those of the 1\textdegree\ and 2\textdegree\ veins (see Tab.~\ref{table:dataset_annotation_time}). It is more practical to prepare partially labeled data with only annotations of 1\textdegree\ and 2\textdegree\ veins when the fully labeled data with annotations of veins of all three orders are unavailable. This has led to an increasing interest in \emph{learning from partially labeled data}. Several attempts have been made towards \emph{partially supervised learning} (PSL) in medical image analysis~\cite{dong2022towards}, where human annotation costs are also high because annotators are required to have clinical expertise. For the first time, we study PSL in the domain of agricultural science. In this paper, we propose a label-efficient learning paradigm with partial supervision to adequately beneﬁt from learning unlabeled, partially labeled, and fully labeled data, simultaneously. This method guides the segmentation of 3\textdegree\ veins based on the existing annotations of 1\textdegree\ and 2\textdegree\ veins. We conduct extensive ablation studies to understand the proposed method and HALVS. The empirical results also provide insights for future research on the task of interest under label scarcity. For example, we notice that the annotation efficiency varies significantly among different species. We are the first to investigate cross-species learning in the leaf vein segmentation task, where the segmentation model is trained by leaf venation data of species that are relatively more amenable to be annotated and makes predictions on the other species whose venation data are challenging to be annotated. The experiments show that cross-species learning is a challenging task under label scarcity and the annotations in HALVS can benefit future studies.
 
We summarize our contributions as follows:
\begin{itemize}
\item To the best of our knowledge, we release the \textbf{first} hierarchical leaf venation segmentation dataset, namely, HALVS, thereby establishing a benchmark for subsequent work and benefiting future studies.
\item This is the first study that leverages partially labeled data in the plant phenomics domain and presents a new practical task domain for partially supervised learning.
\item We evaluate the proposed method extensively on HALVS and achieve superior performance against state-of-the-art baselines. The experimental results also present initial empirical findings on hierarchical vein segmentation, paving the way for future investigations on the task of interest in the context of data scarcity.
\end{itemize}

\section{Related Work}
\subsection{Leaf Vein Datasets}
\label{sec:leaf vein dataset}
Due to the high human cost of pixel-level annotation, there are only a few public leaf vein datasets. To our best knowledge, none of these datasets offer annotations for leaf veins with hierarchical orders. To differentiate HALVS from the existing datasets, the limitations of these datasets are summarized in Tab.~\ref{table:dataset_descirption} in the aspects of image clarity, completeness of leaf blade, whether the raw leaf is chemically treated, and hierarchy of vein annotations. For example, LVD2021 dataset~\cite{li2022leaf} contains 4,977 low-contrast and low-resolution images captured by smartphone cameras and pixel-wise annotations do not provide detailed structural information. Additionally, only 20 high-quality images have been annotated with non-human techniques using histogram equalization and binarization algorithms in~\cite{iwamasa2023network}.
HALVS is the first large-scale annotated dataset for high-quality images.

\begin{table}[]
\centering
\setlength{\tabcolsep}{0.5pt}
\captionsetup{justification=justified,width=\columnwidth, labelfont=small, font=footnotesize}
\begin{tabularx}{0.47\textwidth}{@{\extracolsep{\fill}}c>{\small}c>{\small}c>{\small}c>{\small}c>{\small}c@{\extracolsep{\fill}}}
\toprule
{\scriptsize {Dataset}} & {\scriptsize \textbf{Hier. Annot. }} & {\scriptsize{Untreated}} & {\scriptsize{Complete}} & {\scriptsize{Clear}} & {\scriptsize {\# Images}}\\
\midrule
\makecell[c]{{\scriptsize Leaf Vein Dataset}\\{\scriptsize \cite{blonder2019leaf}}} &  &  &   &  \checkmark &  726\\
\makecell[c]{{\scriptsize LVD2021}\\{\scriptsize \cite{li2022leaf}}} &  &  \checkmark & \checkmark &  &  4,977\\ 
\makecell[c]{{\scriptsize Cleared Leaf Database} \\ {\scriptsize \cite{iwamasa2023network}}} &   &  & \checkmark & \checkmark & 328 \\
\makecell[c]{{\scriptsize Untreated Leaf Dataset} \\ {\scriptsize \cite{iwamasa2023network}}} &  & \checkmark & \checkmark & \checkmark & 479 \\
\midrule  
{\scriptsize HALVS~(Ours)} & \checkmark & \checkmark & \checkmark &\checkmark & \textbf{5,057}\\ 
\bottomrule
\end{tabularx}
\caption{Comparison of HALVS with existing leaf vein segmentation datasets.  ``Hier. Annot.'' means providing annotations of 1\textdegree, 2\textdegree, and 3\textdegree\ veins. ``Untreated'' indicates that the leaf is not chemically treated before imaging. ``Complete'' indicates providing the complete leaf blade. ``Clear'' refers to an image clearly displaying the detailed structural information of the 3\textdegree\ veins.}
\label{table:dataset_descirption}
\end{table}

\subsection{Partially Supervised Learning}
An important research field of PSL is medical image analysis, where partially labeled datasets of different classes of interest are collected from different data sources due to high annotation cost~\cite{dong2022revisiting}. Many efforts have been paid into partially supervised medical image segmentation~\cite{zhou2019prior,fang2020multi,shi2021marginal,dong2022towards}, a similar yet different task to leaf vein segmentation. First, in contrast to medical image segmentation, leaf vein segmentation is still an under-explored task. For medical images, transfer learning has been proven as an efficient solution for cross-sites and cross-modality setups, because of human structural similarity~\cite{dong2022towards}. However, this remains unclear in the leaf vein segmentation. Second, a direct application of PSL methods designed for medical images is infeasible for leaf images because of different image characteristics and task setups. For example, compared with leaf images in HALVS, medical images in previous PSL studies are ``low-quality'' in terms of resolution. Besides, the pixels of leaf veins are scarce, in contrast to most organs or human structures. The desired algorithms should be sensitive to pixel-level information. In this work, we propose the first PSL solution to leaf vein segmentation.

\section{The HALVS Dataset}
    \label{sec:HALVS_dataset}
\subsection{Data Collection}
We collect 5,057 high-quality leaf images with precise venation details from three characteristic species to represent distinct venation characters, including 2,610 images of soybean, 1,947 images of sweet cherry, and 500 images of London planetree  (see Fig.~\ref{fig:annotation_description}). The leaf images are acquired using a simple but efficient transmission scanning method, proposed by~\cite{gan2019automatic}. After being freshly picked from the tree and simply wiped clean, leaves are placed in the flat-bed scanner (Epson Perfection V850 pro scanner) with the transmission scan mode. Images with 4800 × 6000 pixels at a resolution of 600 DPI are captured and saved in the 48-bit RGB PNG format. The leaf petiole can interfere with the segmentation of the primary vein, so in each leaf image, we crop it out and only retain the leaf blade. Then, we apply the minimum bounding rectangle algorithm to eliminate redundant background pixels and generate the final leaf image data. 
\begin{figure}[t]
\label{fig:patch_visual}
\captionsetup[subfigure]{labelfont=small, font=footnotesize}
    \centering
    \captionsetup{font=footnotesize, width=\columnwidth, justification=justified}
    \subcaptionbox{Soybean\label{Orginal image}}[0.15\textwidth]
    {
        \includegraphics[width=0.13\textwidth]{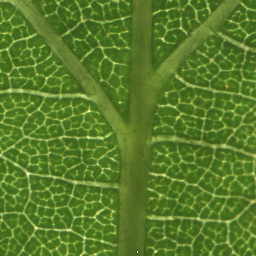}
        \includegraphics[width=0.13\textwidth]{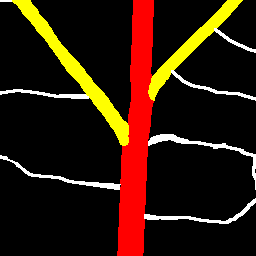}
    }
    \subcaptionbox{Sweet cherry\label{Orginal image2}}[0.15\textwidth]
    {
        \includegraphics[width=0.13\textwidth]{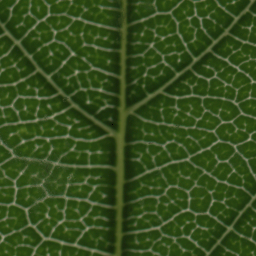}
        \includegraphics[width=0.13\textwidth]{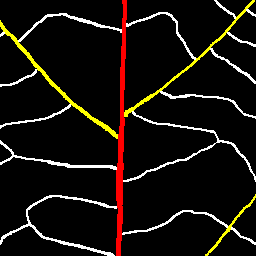}
    }
    \subcaptionbox{London planetree\label{Orginal image3}}[0.15\textwidth]
    {
        \includegraphics[width=0.13\textwidth]{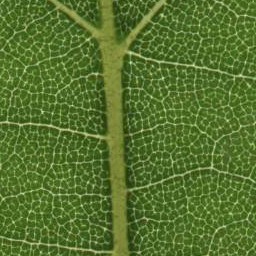}
        \includegraphics[width=0.13\textwidth]{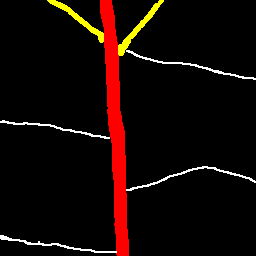}
    }
    \caption{Visualizations of leaves and corresponding vein annotations in HALVS (red: 1\textdegree, yellow: 2\textdegree, white 3\textdegree). Leaf patches are cropped in size of $256\times256$ for illustration purposes.}
    \label{fig:patch}
\end{figure}
\subsection{Classes and Annotations}
Determining the categories, or \emph{orders}, of veins is the first and the most important step during the process of vein annotations. We recognize 1\textdegree, 2\textdegree, and 3\textdegree\ veins starting at the widest 1\textdegree\ vein and progressing to the finest 3\textdegree\ veins, following the roles and vein characteristics described in~\cite{hickey1973classification} and~\cite{ellis2009manual}. Generally, it is relatively easier to recognize the 1\textdegree\ and 3\textdegree\ veins, but sometimes the 2\textdegree\ veins are more complex as they may comprise several subsets with varying widths and courses. Nevertheless, all the vein subsets located between the 1\textdegree\ and 3\textdegree\ veins are considered to be 2\textdegree\ veins. Fig.~\ref{fig:patch} displays examples of our annotations.
\begin{figure*}[t!]
  \centering
  \captionsetup{justification=justified, font=footnotesize}
  \includegraphics[width=0.95\textwidth]{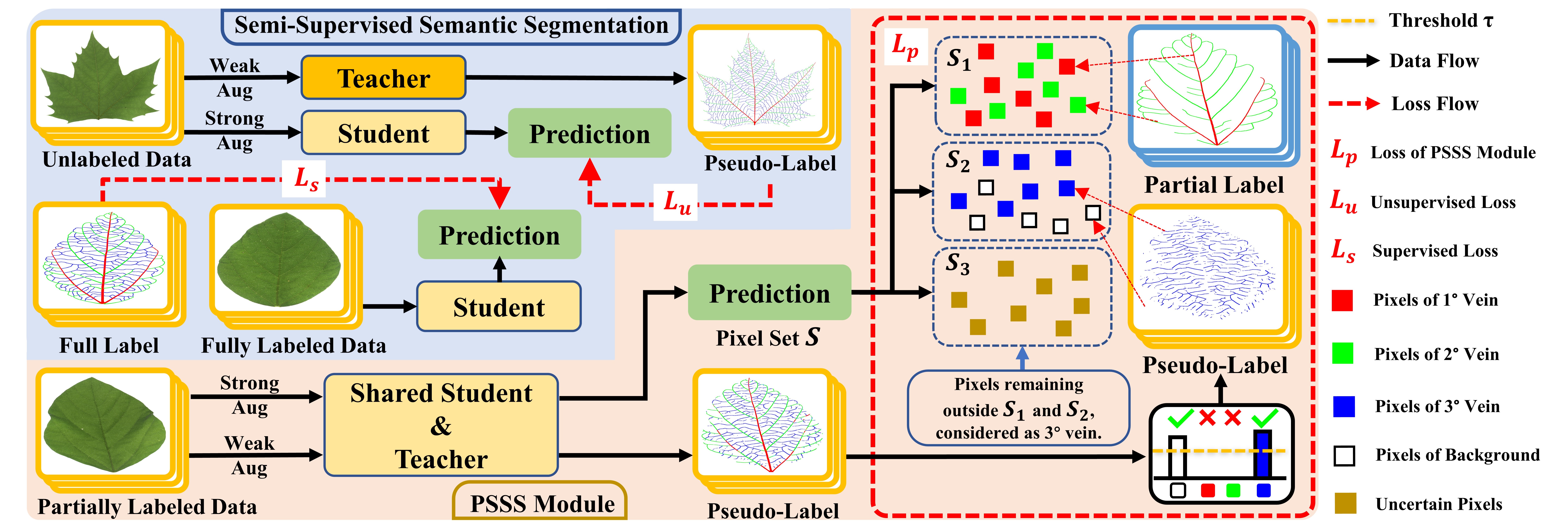}
\caption{Illustration of the proposed label-efficient learning framework for hierarchical leaf vein segmentation. The input of this framework includes unlabeled, partially labeled (\ie~1\textdegree\ and 2\textdegree\ vein), and fully labeled  (\ie~background, 1\textdegree, 2\textdegree, and 3\textdegree\ veins) data which are partitioned into leaf patches with a size of 256 × 256 pixels. Intuitively, complete leaf images instead of patches are used for illustration purposes. The labeled and unlabeled data are handled by a semi-supervised semantic segmentation method. The proposed partially supervised semantic segmentation (PSSS) module can be integrated with any semi-supervised learning framework to handle the partially labeled data.  Here, a teacher-student model~\protect\cite{wang2022semi} is depicted as an example (blue region). In PSSS (orange region), the teacher model generates pseudo-labels from the weakly augmented data, while the student model generates set of predictions $\mathcal{S}$ from the strongly augmented data. $\mathcal{S}_1$ contains the pixels that can be directly supervised by the ground truth of 1\textdegree\ and 2\textdegree\ veins from the partial labels. $\mathcal{S}_2$ contains the pixels that are predicted as pseudo-labels of background and 3\textdegree\ vein with high confidence. All remaining pixels are considered as 3\textdegree\ vein. Three sets of pixels are trained with three different losses. }
  \label{fig:main_fig}
\end{figure*}
For each image, all three orders of veins are traced manually at full width. Despite the efforts, the annotation process for the HALVS dataset remains notably time-consuming. As detailed in Tab.~\ref{table:dataset_annotation_time}, the average time for fully labeling the three orders of veins in a leaf blade image is about 2.1, 3.5, and 7.8 hours for soybean, sweet cherry, and London planetree, respectively. The significantly longer annotation time for the London planetree compared to the other two species is due to the larger leaf size. Labeling the 3\textdegree\ veins consumes considerably more time than labeling the 1\textdegree\ and 2\textdegree\ veins. As a result, 150 images (50 images per species) are densely annotated with fine-grained semantic segmentation labels, \ie~7,831,792 pixels for 1\textdegree\ veins, 10,459,325 pixels for 2\textdegree\ veins, and 16,075,087 pixels for 3\textdegree\ veins. This is equivalent to around 13,000 non-overlapping patches in a size of $256 \times 256$. The whole annotation process takes up to 83.8 person-days by four experienced annotators following biological instructions.

\section{Method}
\subsection{Problem Definition}

In the task setup of interest, the dataset $\mathcal{D} = \mathcal{D}_l \cup \mathcal{D}_u \cup \mathcal{D}_p$ consists of fully labeled, unlabeled, and partially labeled sets. $\mathcal{D}_l = \{(x_i, y_i)\}_{i=0}^{N_l}$ is the fully labeled set, where $x_i$ represents an image and each pixel belongs to one of four classes (\ie~the background, 1\textdegree, 2\textdegree, and 3\textdegree\ veins). The corresponding $y_i$ provides pixel-wise labels for $x_i$.
$\mathcal{D}_u = \{x_j\}_{j=0}^{N_u}$ is the unlabeled set.
$\mathcal{D}_p = \{(x_k, y_k)\}_{k=0}^{N_p}$ is the partially labeled set where $y_k$ only contains label information regarding 1\textdegree\ and 2\textdegree\ veins, missing the background and 3\textdegree\ veins, due to extremely high annotation cost of 3\textdegree\ vein. That is to say, we are only certain whether the pixel belongs to either 1\textdegree\ or 2\textdegree\ veins. 

As shown in Tab.~\ref{table:dataset_annotation_time}, annotating veins are time-consuming and expertise-demanding, and particularly, obtaining 3\textdegree\ vein labels is more challenging. 
We simulate a practical situation: $\mathcal{D}_l$ and $\mathcal{D}_p$ are \textbf{small-scale} datasets and $\mathcal{D}_u$ is a \textbf{large-scale} dataset, \ie~$N_l \ll N_u$ and $N_p \ll N_u$.
The learning objective is to train a leaf vein segmentation model with $\mathcal{D}$. In contrast to standard SSL, we conjecture that the existence of $\mathcal{D}_p$ plays an important role in enhancing the model performance across all four classes.

\subsection{Partially Supervised Semantic Segmentation}
\label{sec:Overall}
We propose a pluggable module for partially supervised semantic segmentation, which we denote as PSSS in the following context. The PSSS module can be easily integrated with state-of-the-art SSL frameworks~\cite{wang2022semi,yang2023revisiting}. The overall framework is illustrated in Fig.~\ref{fig:main_fig}. The training on $\mathcal{D}_l$ and $\mathcal{D}_u$ follows standard SSL methods, where we use $L_S$ and $L_U$ to denote the supervised loss and unsupervised loss, respectively.

For simplicity, we assume that $f$ is a neural network and $(x, y) \in \mathcal{D}_p$ is a sample-label pair where $y$ is a \emph{partial} label with respect to 1\textdegree\ and 2\textdegree\ veins. Following FixMatch~\cite{sohn2020fixmatch}, there are a strong augmentation $A_s$ and a weak augmentation $A_w$. We use $p = f(A_s(x))$ as the prediction output and ${\hat{y}} = {\arg\max}(f(A_w(x)))$ as the pseudo-label. Let $\mathcal{S}$ denote the set of pixels. $\mathcal{S}$ can be split into three sets, \ie~$\mathcal{S} = \mathcal{S}_1 \cup \mathcal{S}_2 \cup \mathcal{S}_3$. $\mathcal{S}_1$ consists of the pixels corresponding to the 1\textdegree\ and 2\textdegree\ veins in the partial label $y$, \ie~
\begin{equation}
    \mathcal{S}_1 = \{m | y^m = 1\}_{m \in \mathcal{S}} \cup \{m | y^m = 2\}_{m \in \mathcal{S}},
\end{equation}
where $m$ is the pixel of interest. $\mathcal{S}_{2}$ contains the pixels corresponding to the background and 3\textdegree\ veins with confidence higher than a threshold $\tau$ in the pseudo-label $\hat{y}$, \ie~
\begin{equation}
    \mathcal{S}_2 = \{m | \hat{y}^m > \tau \}_{m \in \mathcal{S}},
    \label{eq:s2}
\end{equation}
where $\tau \in (0, 1)$ is the hyperparameter. 
$\mathcal{S}_3$ is the complement of $\mathcal{S}_1 \cup \mathcal{S}_2$ under $\mathcal{S}$.

For $\mathcal{S}_1$, the ground truth labels are available. The standard supervised learning is applied.
\begin{equation}
L_{P}^{s} = \frac{1}{|\mathcal{S}_1|} \sum_{m \in \mathcal{S}_1} H(y^{m}, f(A_s(x))^{m})
\label{eq:loss_s1}
\end{equation}
In Eq.~\eqref{eq:loss_s1}, $|\cdot|$ denotes the cardinality of the set (\ie~the number of pixels) and $H(\cdot, \cdot)$ denotes the cross-entropy. 

For $\mathcal{S}_2$, an unsupervised loss is computed using the pseudo-label $\hat{y}$. 
\begin{equation}
L_{P}^{u} = \frac{1}{|\mathcal{S}_2|} \sum_{m \in \mathcal{S}_2} H(\hat{y}^{m}, f(A_s(x))^{m})
\label{eq:loss_s2}
\end{equation}
$L_{P}^{u}$ utilizes consistency regularization to improve the model's generalization on unlabeled pixels and mitigate the overfitting. Due to class imbalance, the learned pseudo-labels might be dominated by the majority classes, thus exacerbating the overfitting. Note, in Eq.~\eqref{eq:s2}, only those pseudo-labels with high confidence participate in $L_{P}^{u}$. This excludes unreliable pseudo-labels that might sabotage the training. We will further discuss the impact of $\tau$ in the experiments. 

The pixels in $\mathcal{S}_{3}$ are uncertain pixels, \ie~pixels with missing ground truth labels and with low confidence of the pseudo-labels. As the classes of interest are mutually exclusive, a pixel can only belong to one class. Though the remaining pixels in $\mathcal{S}_{3}$ can be either the background or 3\textdegree\ vein, they are highly possible to be 3\textdegree\ vein. The motivation here is that if the model can reliably predict the pixels belonging to the background, 1\textdegree, and 2\textdegree\ veins, the remaining pixels shall be the 3\textdegree\ vein. We hypothesize that the background class is much easier to predict in contrast to the vein classes. With $L_S$ and $L_U$, the model can efficiently learn the patterns of the background as there are more non-vein pixels than vein pixels. Utilizing this prior knowledge allows the model to focus more on predicting the most challenging 3\textdegree\ vein class, making the segmentation task straightforward. We design a class-specific exclusion loss~\cite{shi2021marginal} to leverage these uncertain pixels:
\begin{equation}
L_{P}^{c} = \frac{1}{|\mathcal{S}_{3}|} \sum_{m \in \mathcal{S}_{3}} e \cdot \log(1 + f(A_s(x))^{m}),
\label{eq:loss_s3}
\end{equation}
where $e = [1,1,1,0]$. As the labels for the 3\textdegree\ vein is scarce, $L_{P}^{c}$ also regularizes the class imbalance.

To summarize, the loss for $\mathcal{D}_p$ is the sum of the three component losses above, \ie~$L_P = L_{P}^s + L_{P}^{u} + L_{P}^{c}$.
And the total loss for the optimization process is $L = L_S + L_U + \lambda  L_P$,
where $\lambda$ is a hyperparameter controlling the weight of $L_P$.

\begin{figure*}[t]
    \centering
    \captionsetup{font=footnotesize, width=\textwidth, justification=justified}
    \subcaptionbox{Original Image\label{orginal image}}[0.19\textwidth]
    {
        \rotatebox{90}{\includegraphics[height=0.185\textwidth]{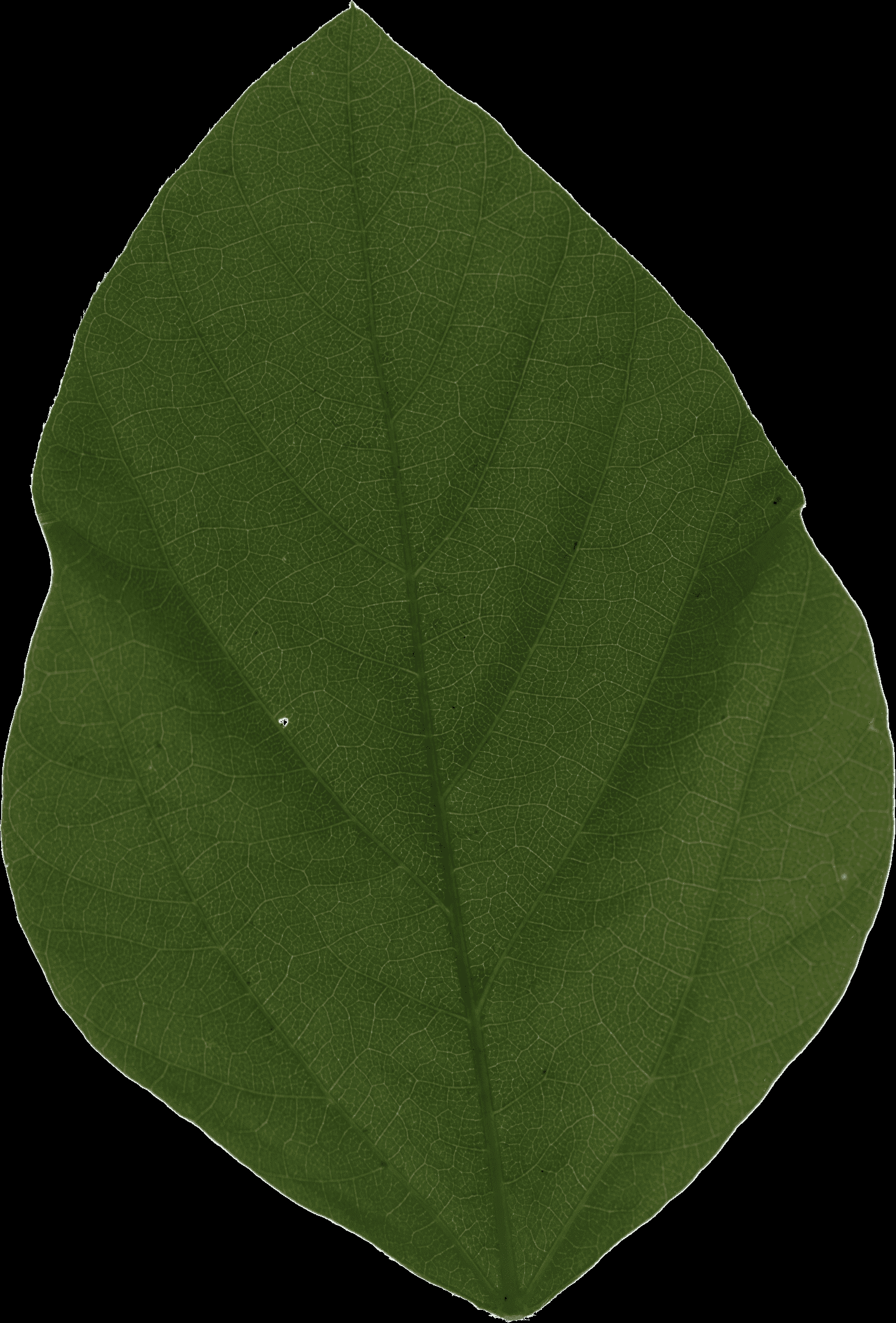}}
        \rotatebox{90}{\includegraphics[height=0.185\textwidth]{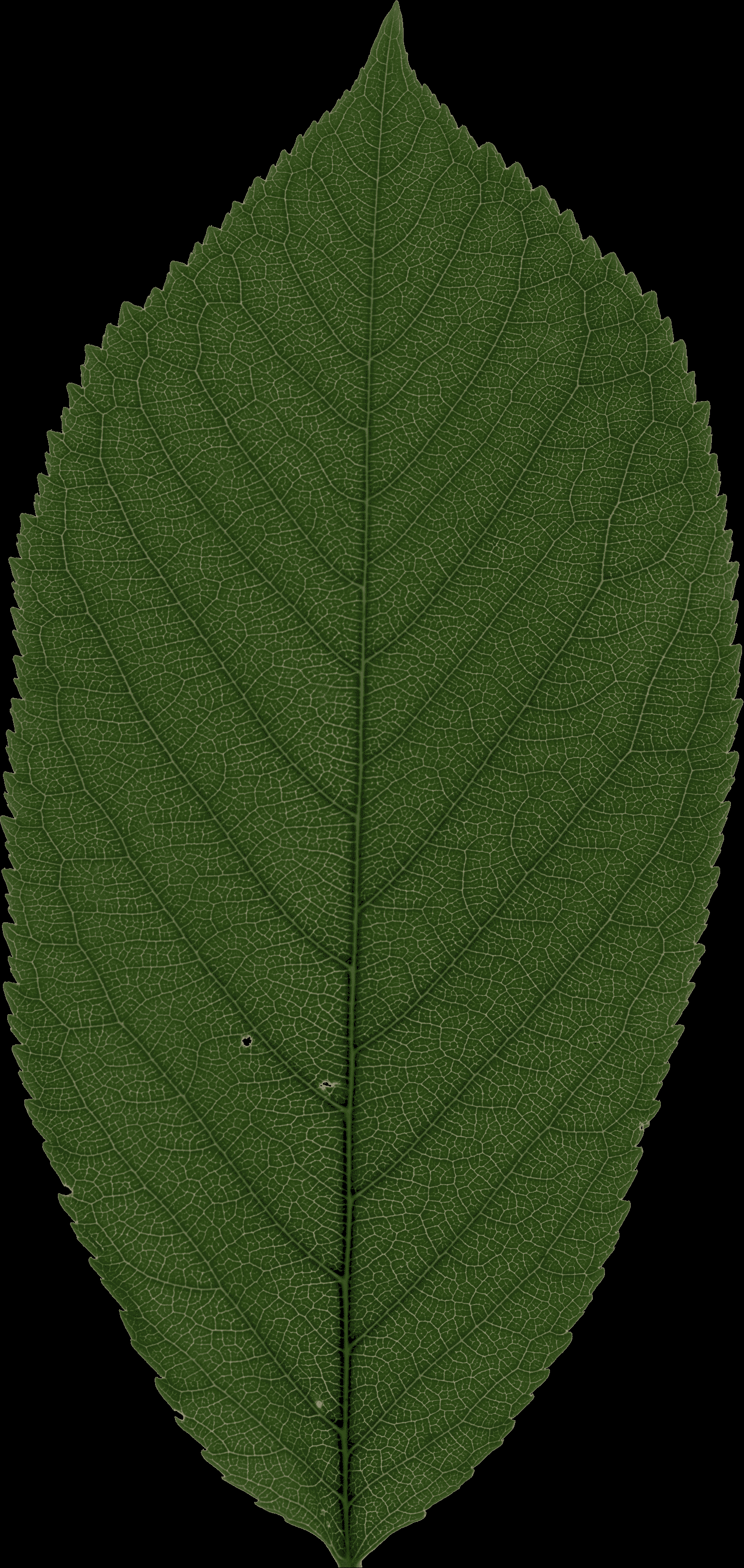}}
        {\includegraphics[width=0.185\textwidth]{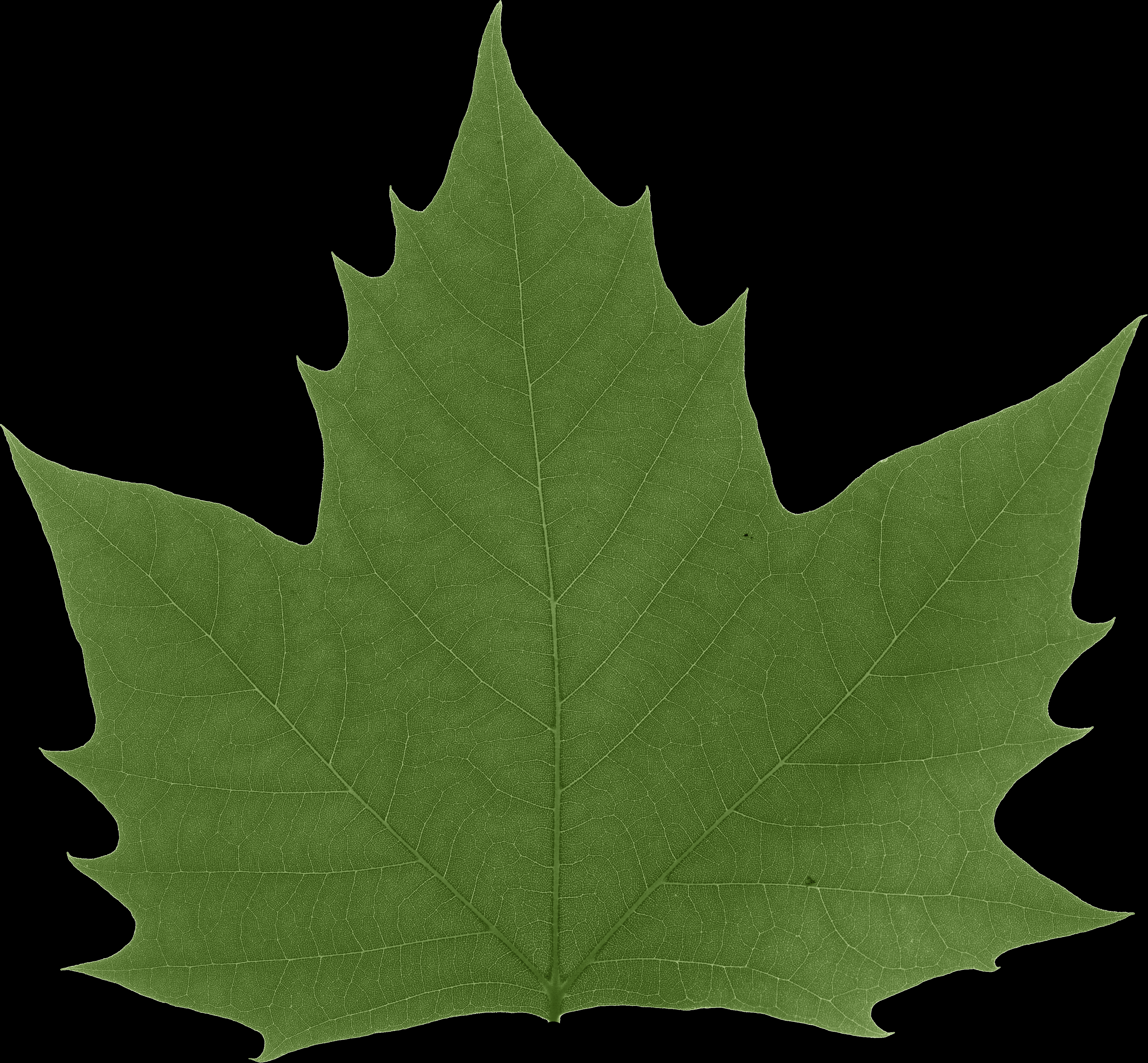}}
    }
    \subcaptionbox{Ground Truth\label{Ground truth}}[0.19\textwidth]
    {
        \rotatebox{90}{\includegraphics[height=0.185\textwidth]{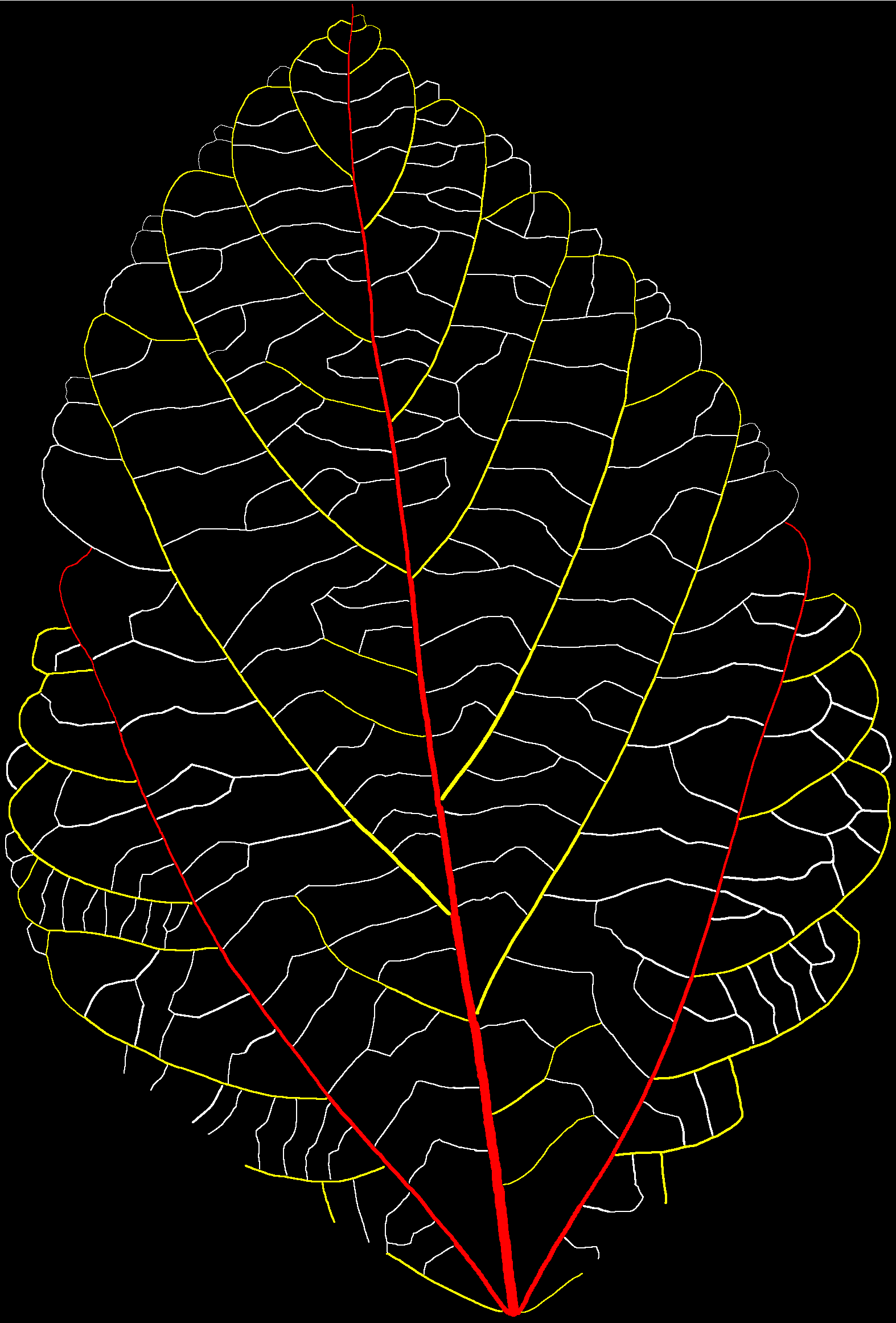}}
        \rotatebox{90}{\includegraphics[height=0.185\textwidth]{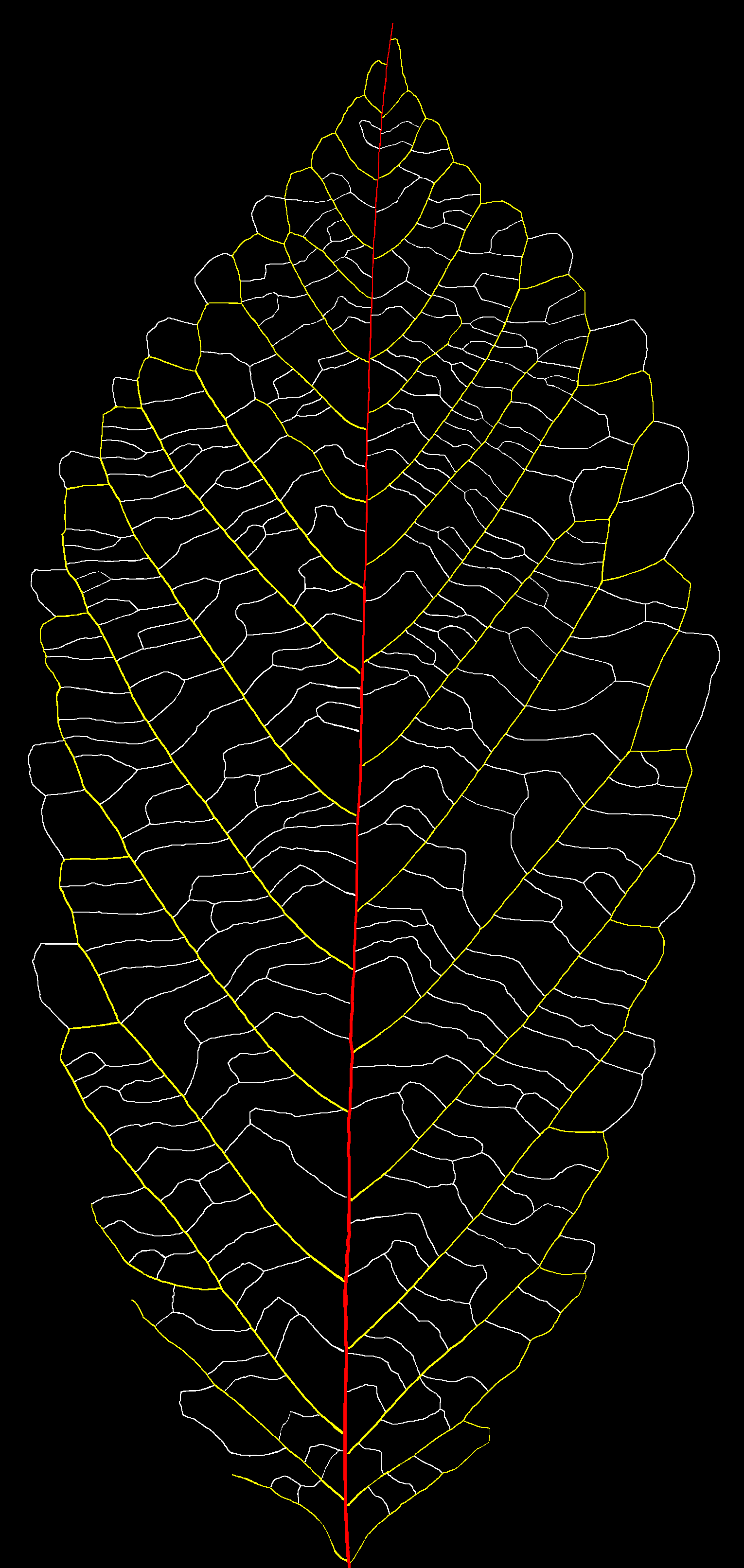}}
        {\includegraphics[width=0.185\textwidth]{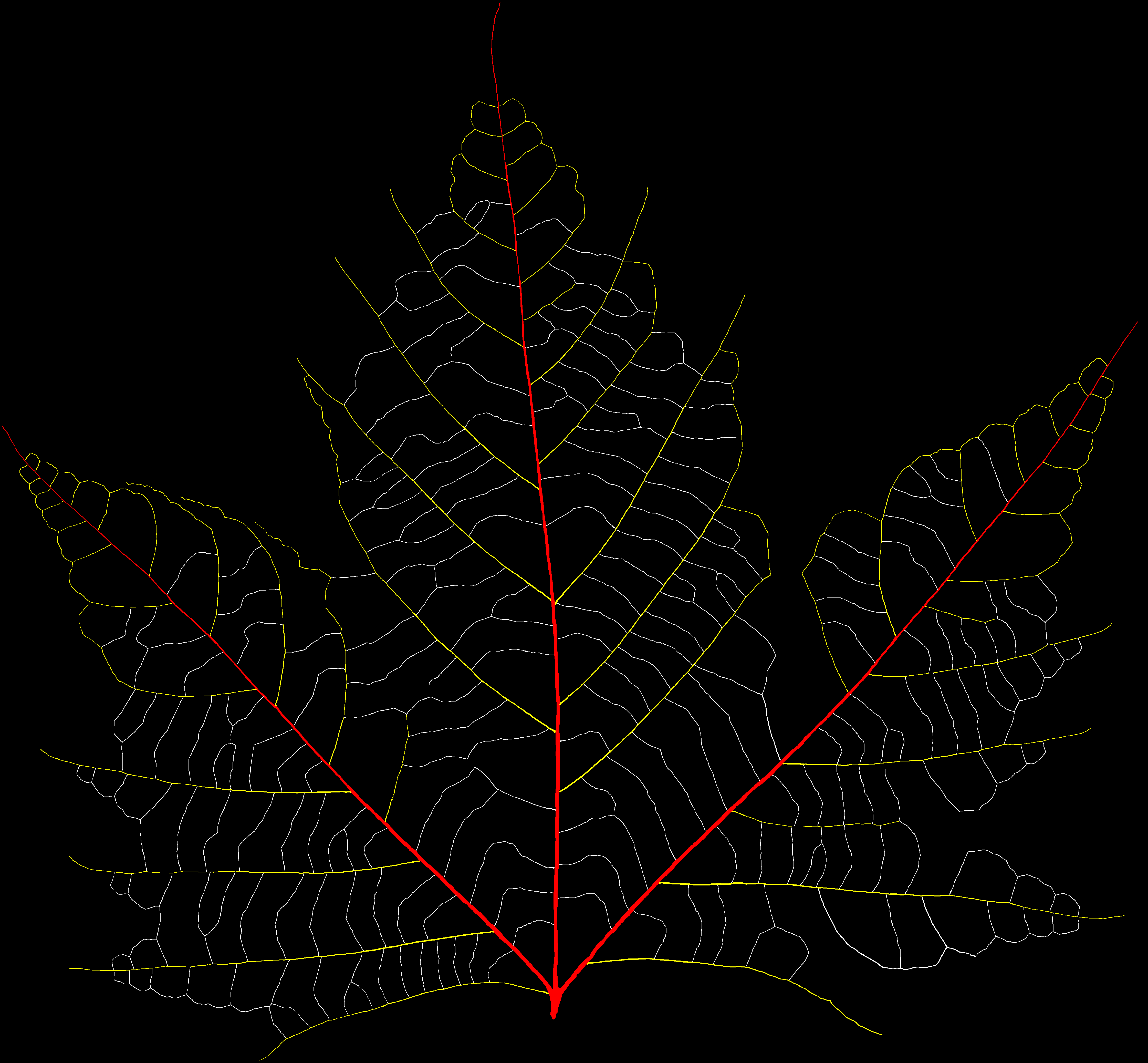}}
    }
    \subcaptionbox{Supervised\label{Supervised (DeepLabv3+)}}[0.19\textwidth]
    {
        \rotatebox{90}{\includegraphics[height=0.185\textwidth]{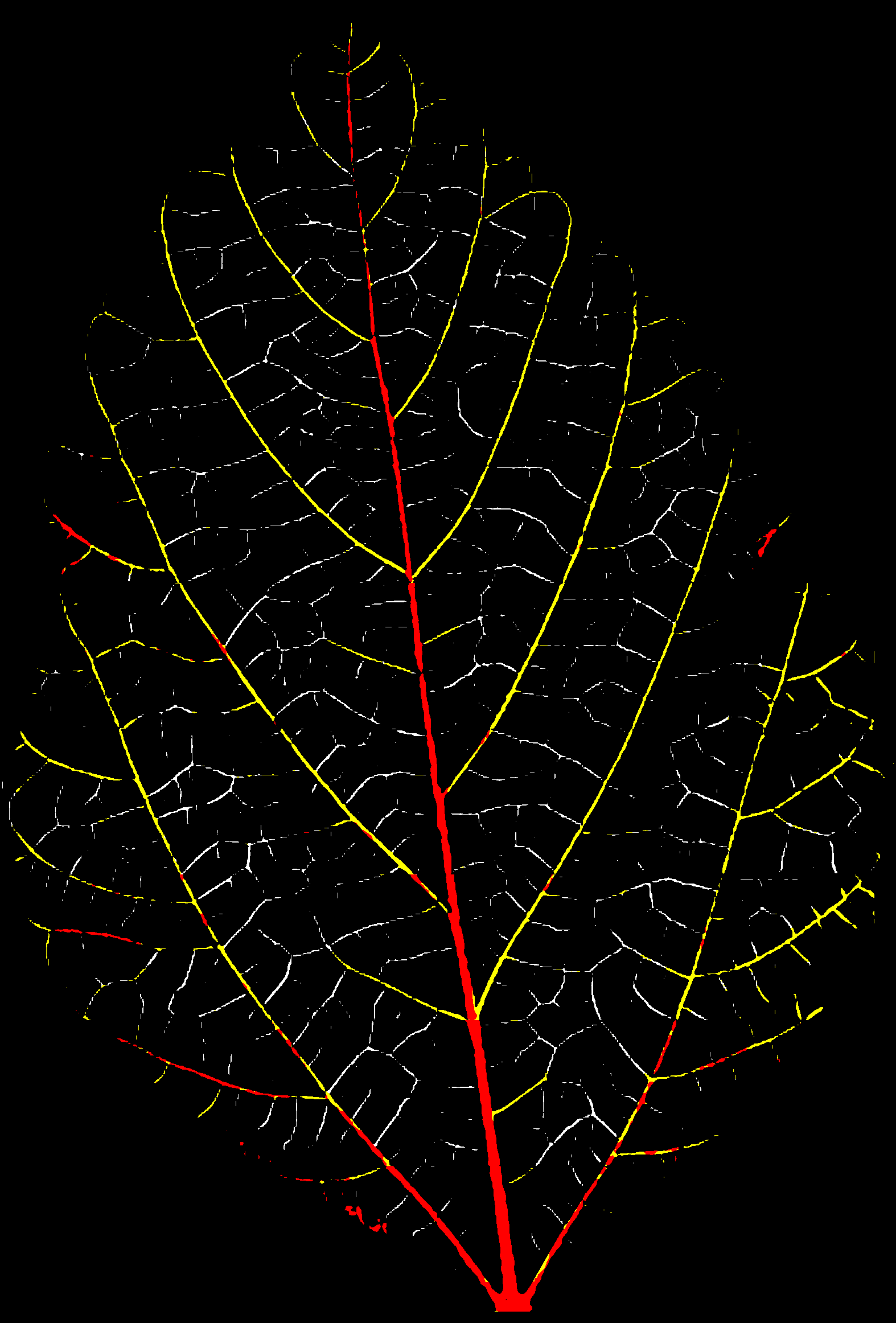}}
        \rotatebox{90}{\includegraphics[height=0.185\textwidth]{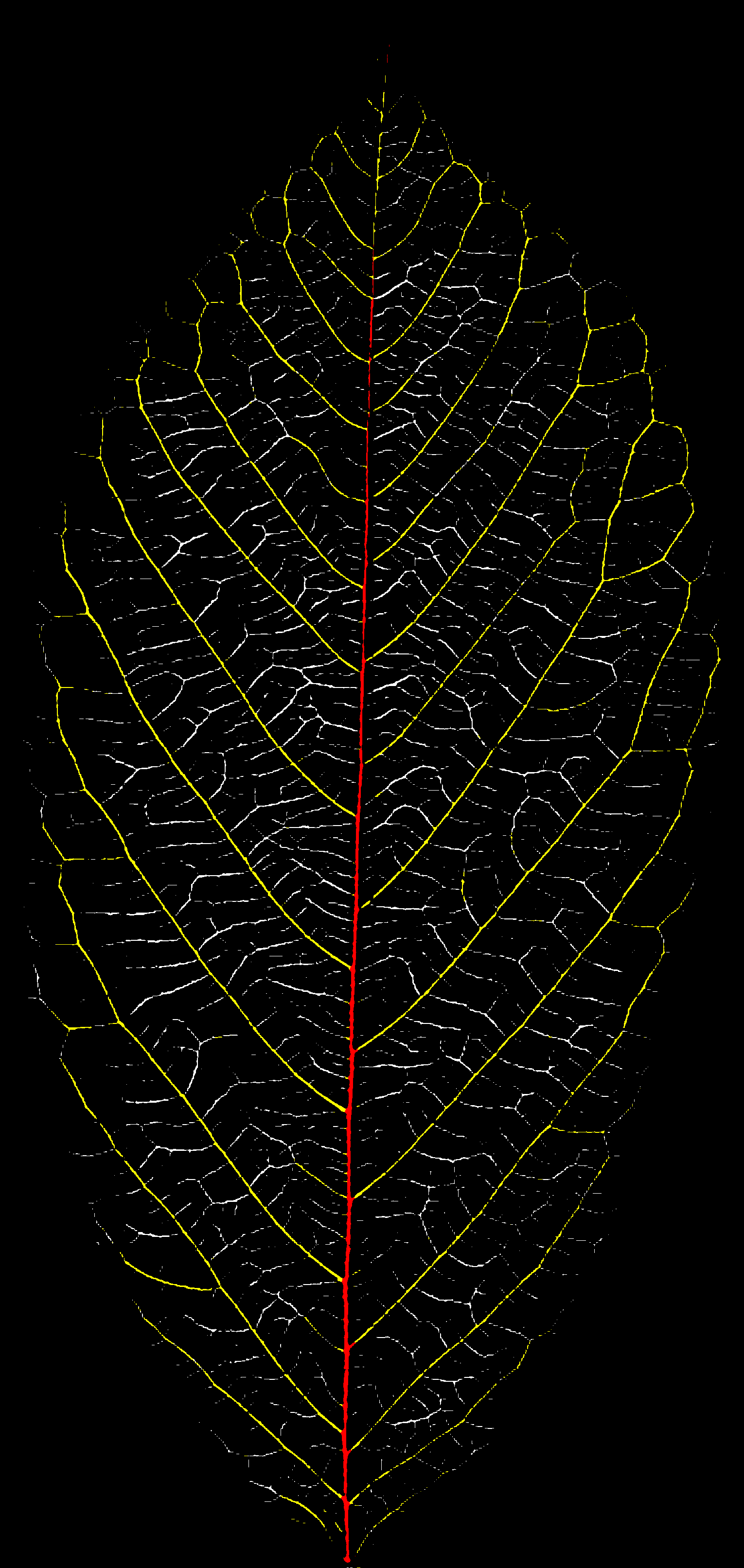}}
        \includegraphics[width=0.185\textwidth]{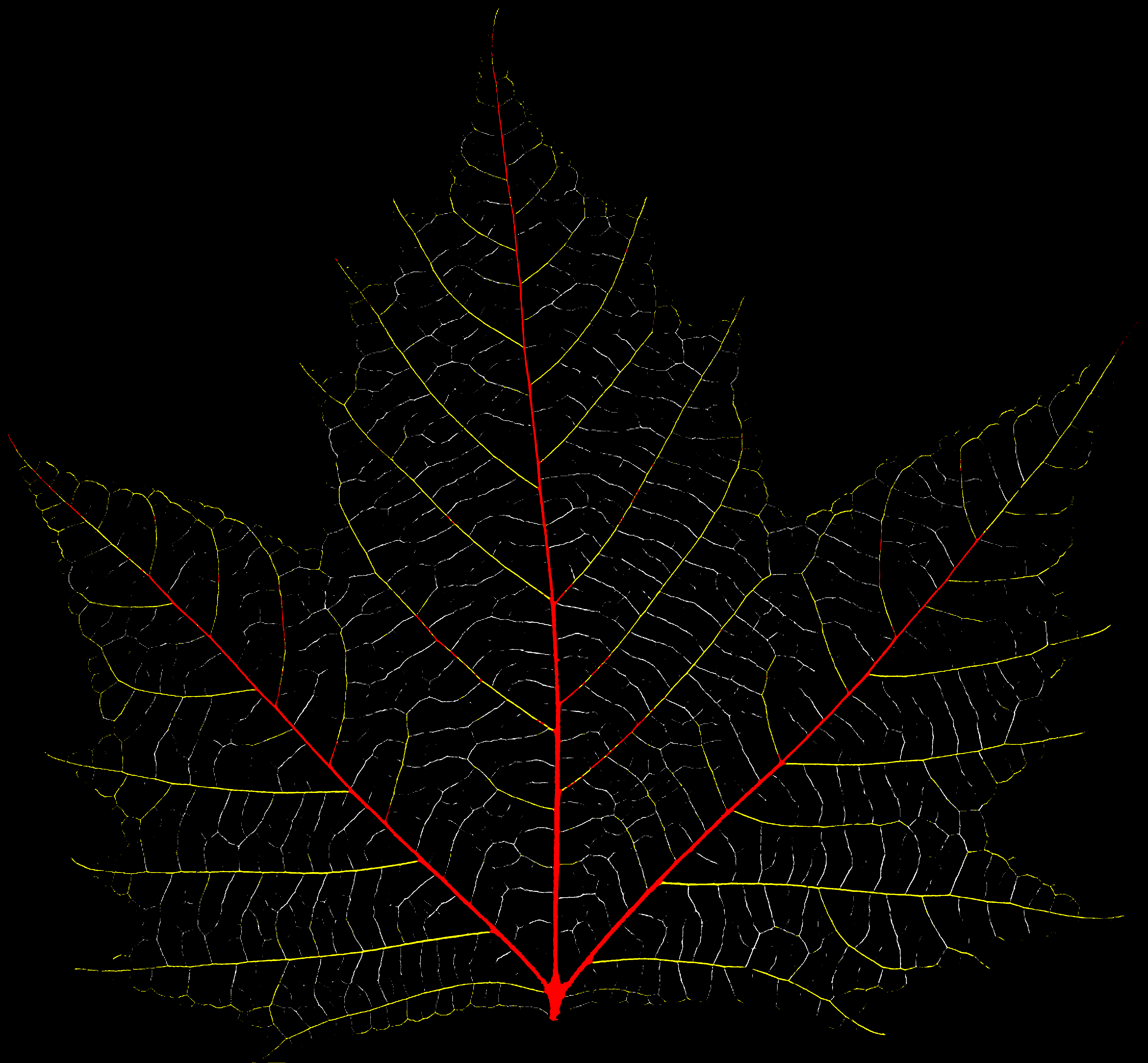}
    }
    \subcaptionbox{Semi-Supervised\label{Semi-supervised }}[0.19\textwidth]
    {
        \rotatebox{90}{\includegraphics[height=0.185\textwidth]{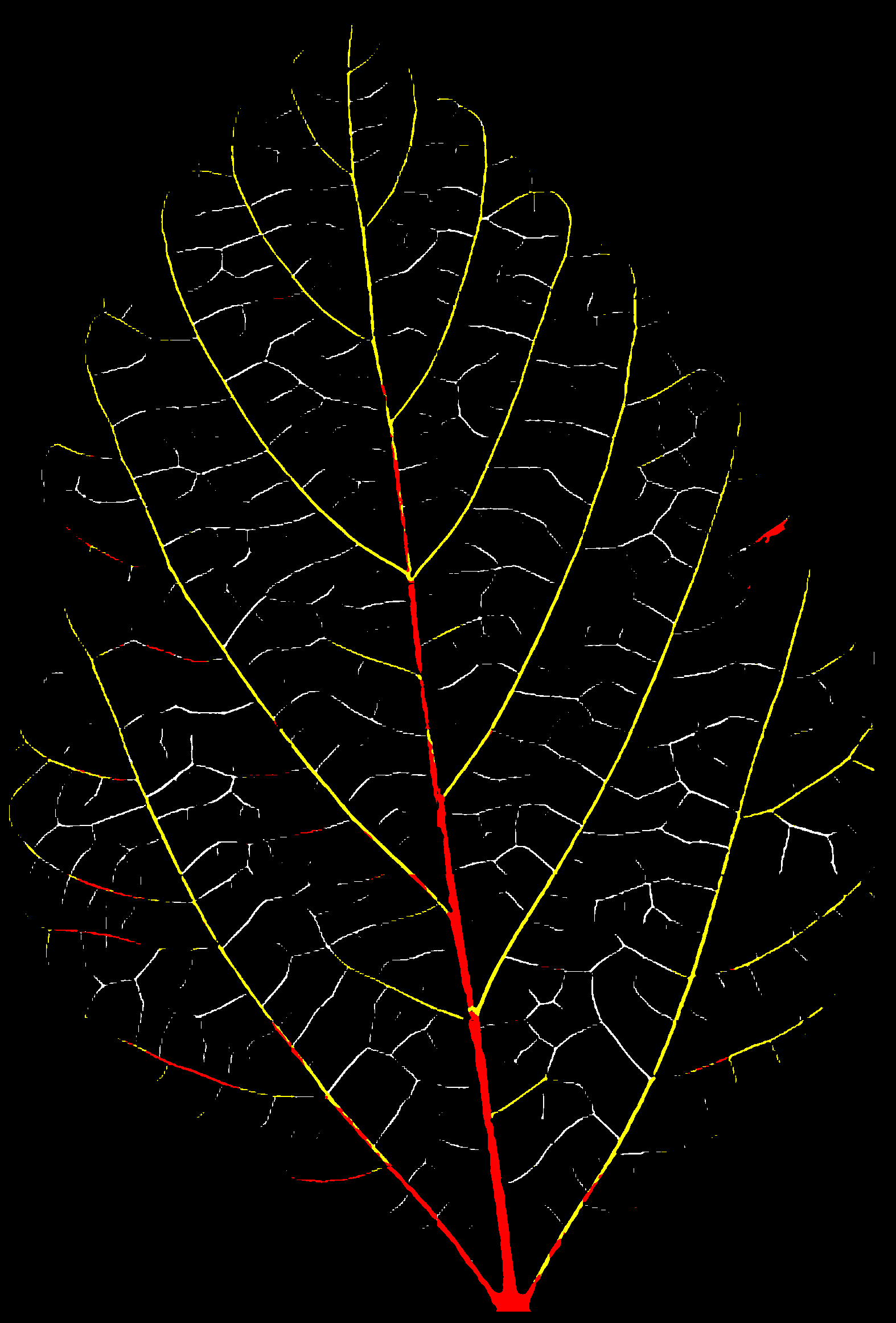}}
        \rotatebox{90}{\includegraphics[height=0.185\textwidth]{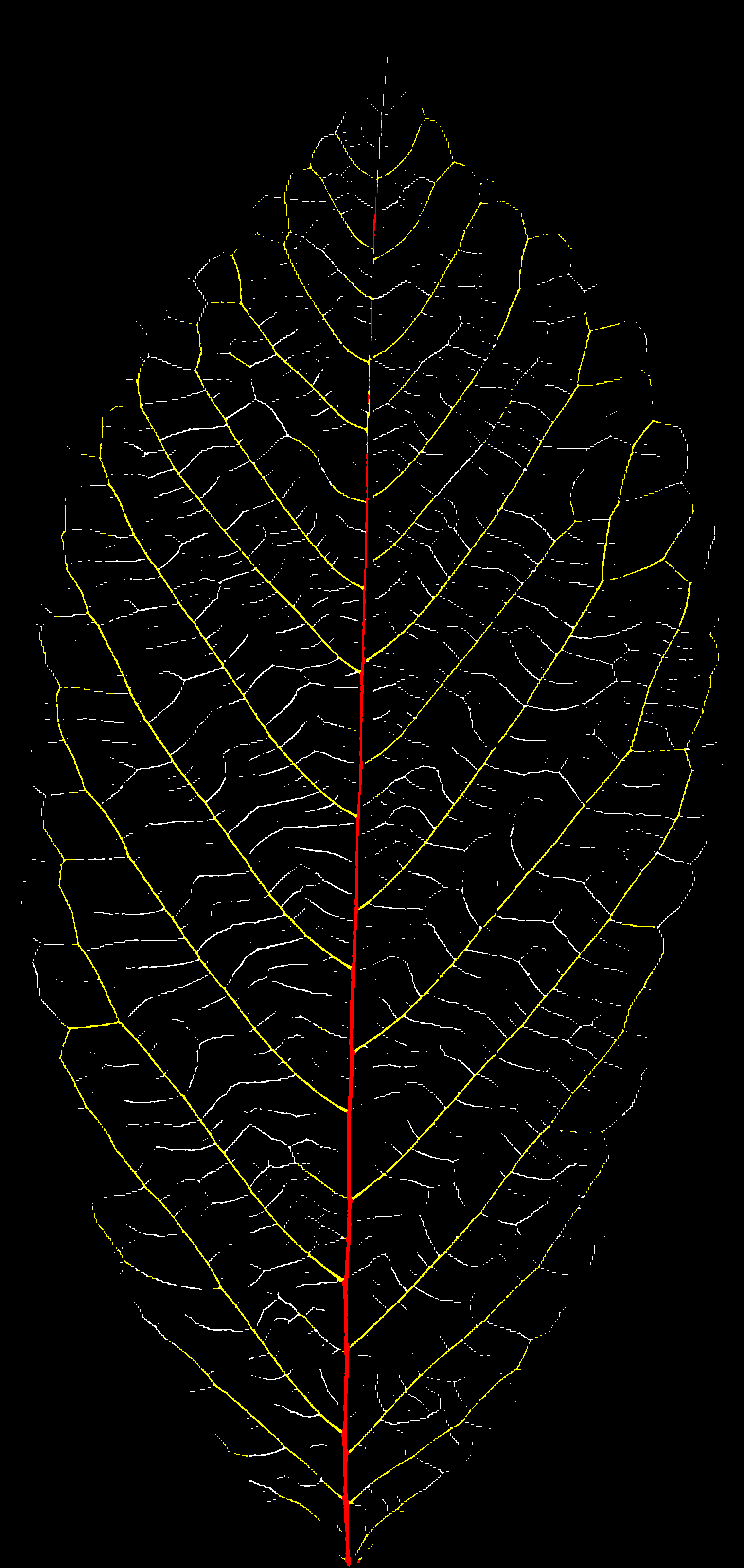}}
        \includegraphics[width=0.185\textwidth]{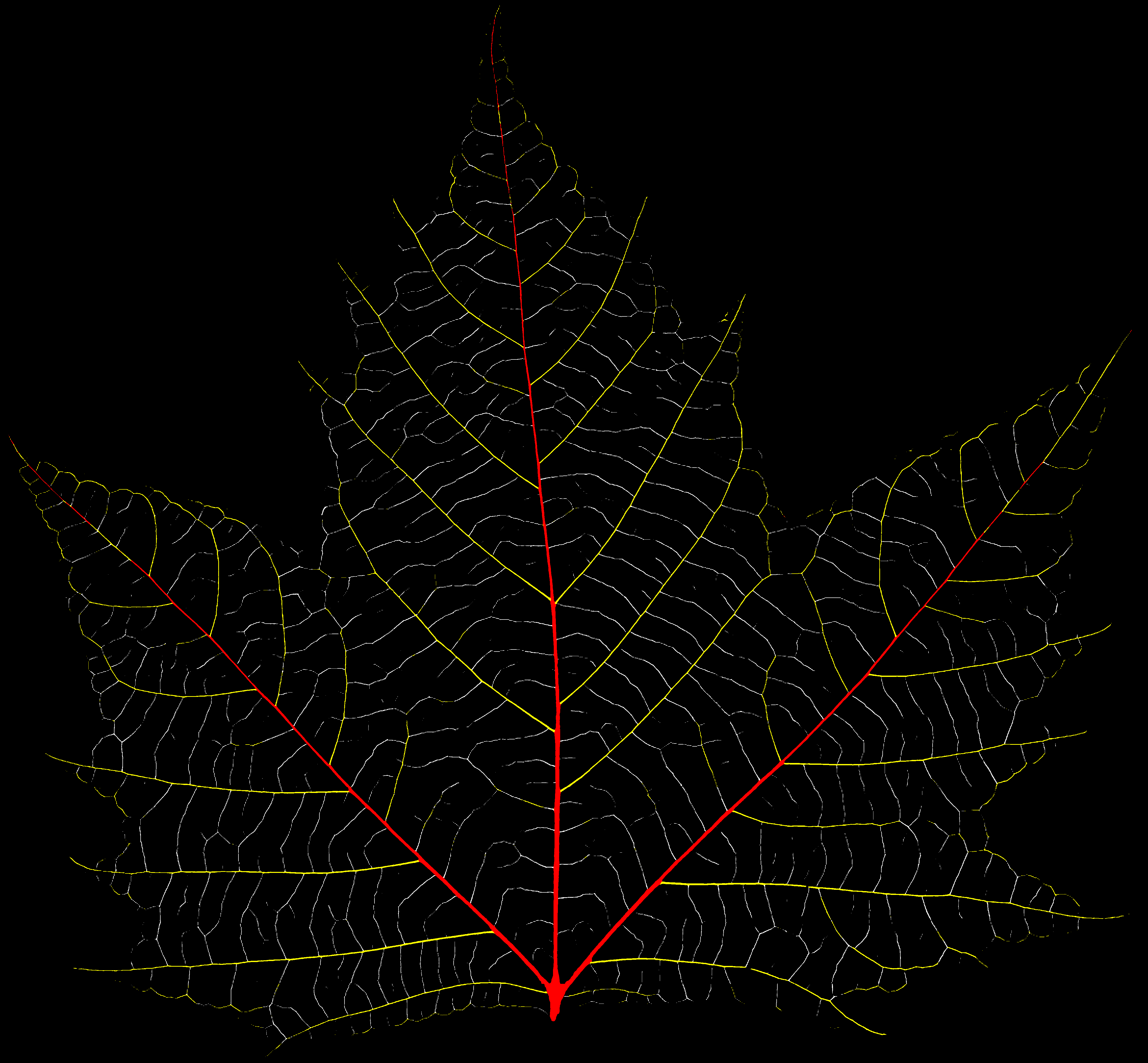}
    }
    \subcaptionbox{Ours }[0.19\textwidth]
    {
        \rotatebox{90}{\includegraphics[height=0.185\textwidth]{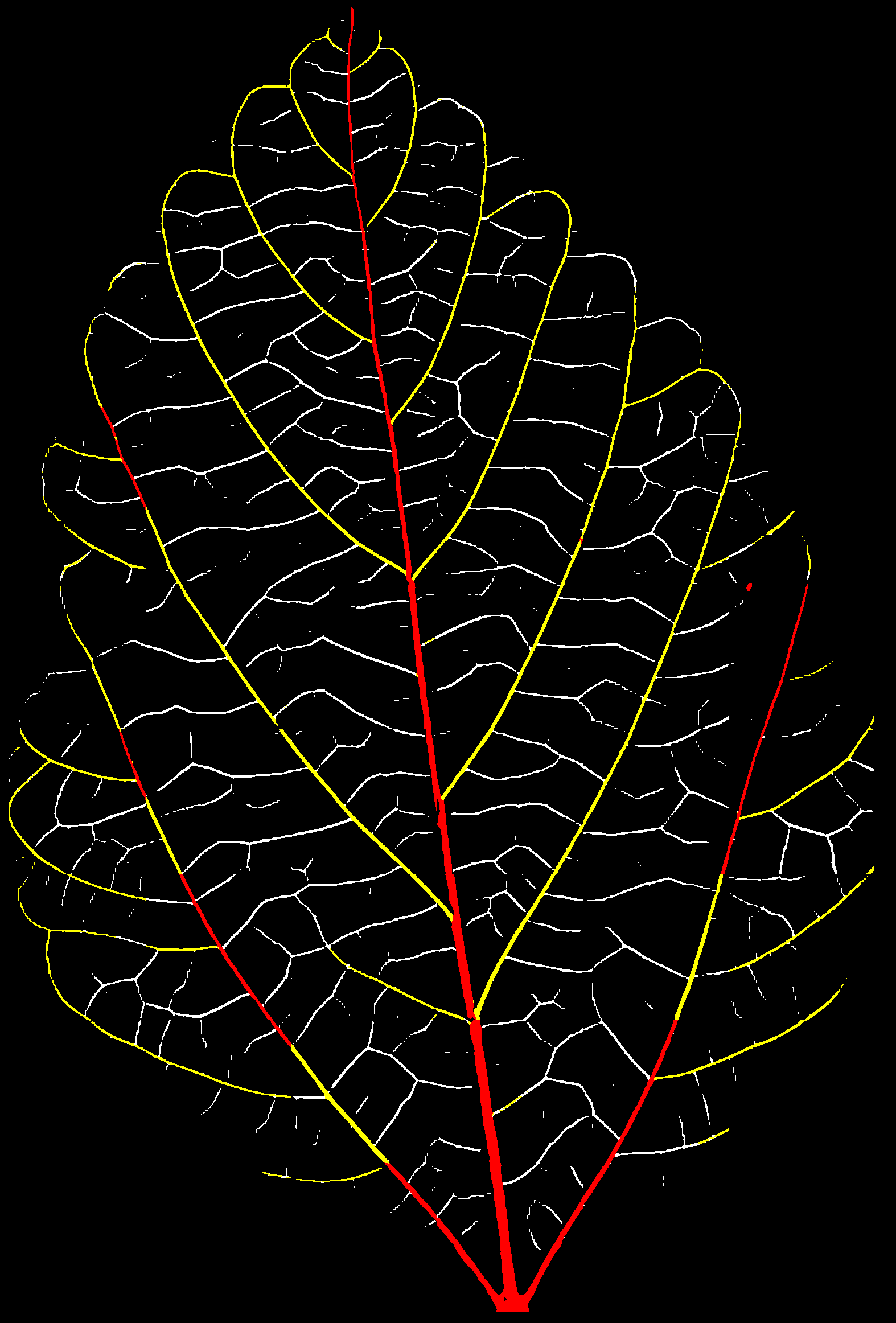}}
        \rotatebox{90}{\includegraphics[height=0.185\textwidth]{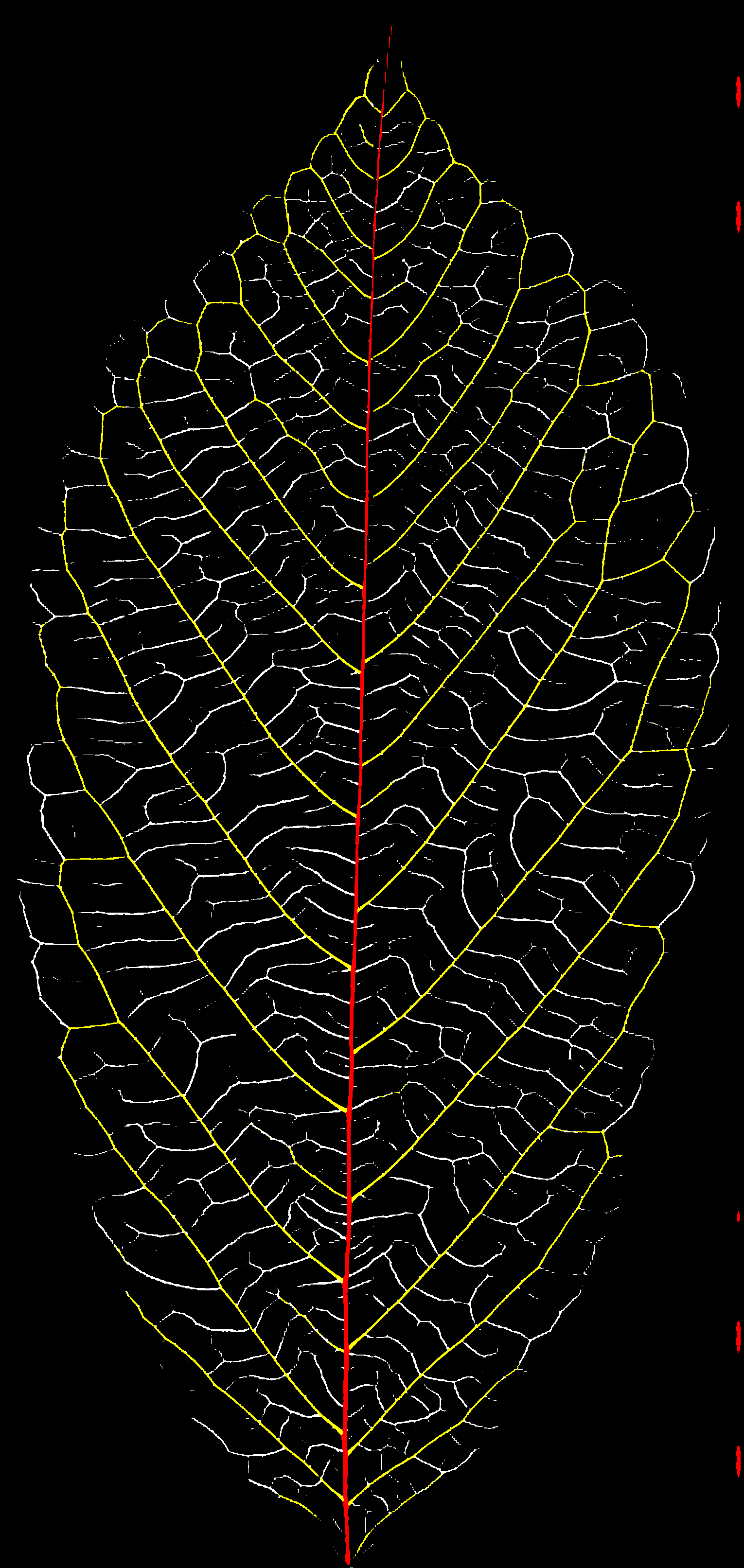}}
        \includegraphics[width=0.185\textwidth]{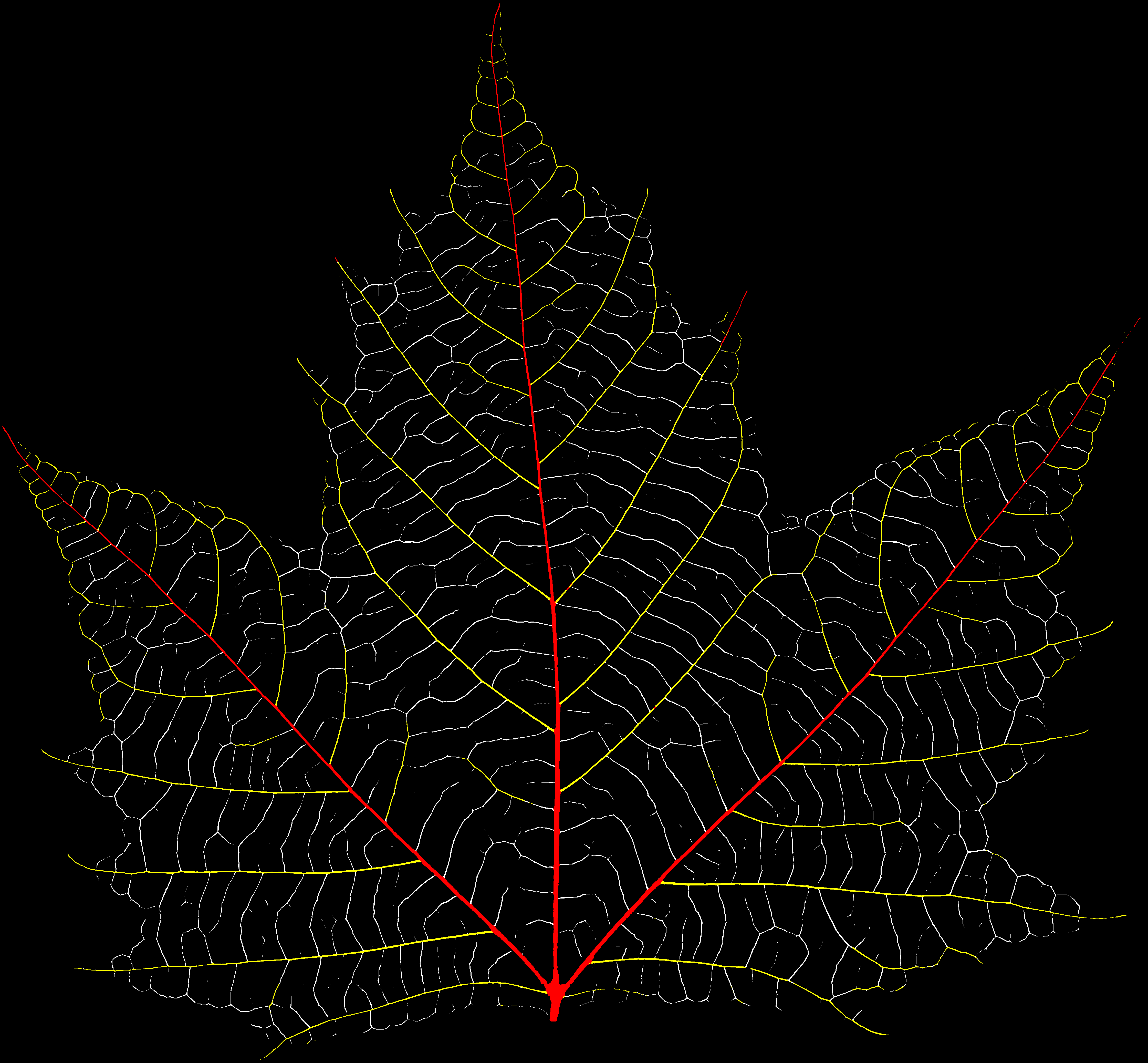}
    }
    \caption{Visualizations of the segmentation results on the HALVS dataset. From top to bottom: soybean, sweet cherry, and London planetree. The colors of red, yellow, and white represent 1\textdegree, 2\textdegree, and 3\textdegree\ veins, respectively. The qualitative performance of the 3\textdegree\ vein in (e) is notably superior to the counterparts in (c) and (d). Best viewed with digital zoom.}
    \label{fig:results}
\end{figure*}
\section{Experiments}
The purpose of experimental design is twofold. First, we aim to evaluate the efficiency of PSSS on HALVS under label scarcity. Second, we aim to leverage HALVS to understand the challenges in cross-species transfer learning.

\subsection{Experimental Setup}
\noindent\textbf{Baselines.}
The PSSS module can easily integrate into semi-supervised semantic segmentation methods without changing the training pipeline, making it convenient for a comprehensive and practical evaluation. We choose three representative approaches as our baselines, namely: FixMatch~\cite{sohn2020fixmatch}, U$^2$PL~\cite{wang2022semi}, and UniMatch~\cite{yang2023revisiting}.
FixMatch is a seminal semi-supervised classification method that combines consistency regularization and pseudo-labeling. We adapt FixMatch to semantic segmentation as a strong baseline. U$^2$PL and UniMatch are two recent state-of-the-art approaches in semi-supervised semantic segmentation. Additionally, we provide an ``Oracle'', where the training data $\mathcal{D}$ is completely labeled. ``Oracle'' is considered as the performance upper bound that a semi-supervised method can achieve. Similarly, we provide a performance lower bound, only using the labeled information to train a supervised model. We denote this baseline as ``Supervised''.

\noindent\textbf{Implementation.}
All experiments are implemented by PyTorch~\cite{paszke2019pytorch} on two NVIDIA 4090 GPUs with 24G memory. The main purpose of the experiments is to evaluate the efficiency of PSSS in the default segmentation network DeepLabv3+~\cite{chen2018encoder} with ResNet-101~\cite{he2016deep} backbone, a strong semantic segmentation architecture. Following U$^2$PL and UniMatch, we use a standard stochastic gradient descent optimizer with a batch size of 4. For a fair comparison, we adopt the poly learning strategy~\cite{chen2017deeplab} for all methods, where the initial learning rate is $10^{-3}$ and the weight decay is $10^{-4}$. We train all methods for 80 epochs and report the best performance measured in mean Intersection-over-Union (mIoU). By default, we set $\lambda = 1$ and $\tau = 0.95$. To retain image details in our high-resolution data, we partition the original images into patches of size $256\times256$. The default data ratio of fully labeled, partially labeled, and unlabeled data is $1:1:10$, where one basic unit denotes 6 images with full leaf blades (around 600 patches). For the class-balance purpose, the default class ratio among three species is $1:1:1$. 

\subsection{Efficiency of PSSS}
The evaluation results of PSSS with three seminal SSL methods are shown in Tab.~\ref{table:main_table}. While the numerical results between different SSL baselines vary, the pattern is clear: adding PSSS to the SSL training can significantly improve the 3\textdegree\ vein IoU by $14.37\%$, $3.89\%$, and $11\%$ on FixMatch, U$^2$PL, and UniMatch, respectively. It also improves the mIoU of FixMatch, U$^{2}$PL, and UniMatch by $8.55\%$, $2.57\%$, and $6.82\%$, respectively.  It is worth mentioning that with PSSS, SSL baselines can achieve competitive performance with ``Oracle'', especially on the most challenging 3\textdegree\ vein.

In addition to quantitative evaluation, we perform qualitative comparison in Fig.~\ref{fig:results}. We use UniMatch as the baseline SSL method and visualize the segmentation results of ``Supervised'', ``UniMatch'' (Semi-Supervised), and ``UniMatch + PSSS'' (Ours)  from the Tab.~\ref{table:main_table}. It can be clearly seen that with PSSS, the segmentation performance is significantly improved across all three categories compared to the supervised and semi-supervised baselines, especially on the 3\textdegree\ vein. Meanwhile, we should also realize that hierarchical leaf vein segmentation is a challenging task and there are still lots of space for improvement.

\begin{table}[t]
\centering
\captionsetup{font=footnotesize, width=\columnwidth, justification=justified}
{\footnotesize
\resizebox{0.47\textwidth}{!}
{
\begin{tabular}{@{\extracolsep{\fill}}lcccc@{\extracolsep{\fill}}}
\toprule
Method & 1\textdegree\ vein & 2\textdegree\ vein & 3\textdegree\ vein & mIoU\\
\midrule
Supervised & 64.93 & 39.71 & 23.56 & 42.73\\
\midrule
\midrule
FixMatch~\cite{sohn2020fixmatch} & 68.07 & 45.57 & 22.32 & 45.32\\
FixMatch + PSSS~(Ours) & {\bfseries72.71} & {\bfseries52.20} & {\bfseries36.69} & {\bfseries53.87}\\
\midrule
\midrule
U$^2$PL~\cite{wang2022semi} & 68.85 & 43.28 & 28.85 & 46.99\\
U$^2$PL + PSSS~(Ours)& {\bfseries69.48} & {\bfseries46.46} & {\bfseries32.74} & {\bfseries49.56}\\
\midrule
\midrule
UniMatch~\cite{yang2023revisiting} & 67.94 & 45.73 & 25.73 & 46.47\\
UniMatch + PSSS~(Ours) & {\bfseries71.79} & {\bfseries51.34} & {\bfseries36.73} & {\bfseries53.29}\\
\midrule
\midrule
Oracle & 73.37  & 51.12  & 33.80 & 52.76 \\
\bottomrule
\end{tabular}%
}

}
\caption{Performance comparison for SSL methods with and without the proposed PSSS module. IoU of 1\textdegree\ vein, 2\textdegree\ vein, 3\textdegree\ vein, and mIoU are reported. Integrating PSSS can efficiently improve the segmentation performance, especially on the 3\textdegree\ vein.}

\label{table:main_table}
\end{table}

\subsection{Ablation Studies}
\label{sec:Ablation_studies}
In this section, we use UniMatch as the baseline SSL method to study the robustness of PSSS under the following setups.

\begin{table}[t]
\centering
\captionsetup{font=footnotesize, justification=justified}
{\footnotesize
\label{tab:my-table}
\begin{tabular}{@{\extracolsep{\fill}}c c c c c c c@{\extracolsep{\fill}}}
\toprule
$L_{P}^{s}$ & $L_{P}^{u}$ & $L_{P}^{c}$ & 1\textdegree\ vein & 2\textdegree\ vein & 3\textdegree\ vein & mIoU\\
\midrule
-& -& -&70.11&47.75&16.73&44.86\\
\checkmark& & &67.53&51.26&18.28&45.69\\
\checkmark & \checkmark& &70.16&57.05&22.08&49.76\\
\checkmark &  &  \checkmark&70.37&54.52&30.50&51.80\\ 
\checkmark& \checkmark& \checkmark&{\bfseries73.27}&{\bfseries57.19}&{\bfseries32.67}&{\bfseries54.38}\\
\bottomrule
\end{tabular}
}
\caption{Ablation study on the effectiveness of loss components in PSSS. In contrast to SSL baseline ($1^{\text{st}}$ row), all loss components have positive impacts on the 3\textdegree\ vein.}
\label{table:loss_ablation}
\end{table}

\noindent\textbf{Effect of Component Losses in PSSS.}
As shown in Tab.~\ref{table:loss_ablation}, all three loss components of PSSS can improve the overall performance. It is interesting that adding our designed $L_{P}^{c}$ not only significantly improves the performance of 3\textdegree\ vein, but also benefits the other two classes ($2^{nd}$ row \emph{vs} $4^{th}$ row, $3^{rd}$ row \emph{vs} $5^{th}$ row). We conclude that  $L_{P}^{c}$ plays an important role in PSSS.

\begin{figure}[t]
\centering
\setlength{\baselineskip}{0pt}
\captionsetup{font=footnotesize, justification=justified}
\begin{subfigure}[t]{\columnwidth}
\centering
\includegraphics[width=0.8\textwidth,height=0.34\textwidth]{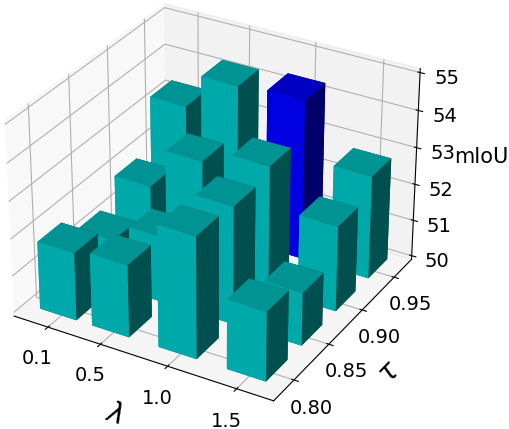}
\label{fig:sub1}
\end{subfigure}
\caption{Comparison of performance under various combinations of score threshold $\tau$ and loss weight $\lambda$. The blue bar represents the optimal situation, with $\lambda = 1$ and $\tau = 0.95$.
}
\label{fig:hyperparameters}
\end{figure}

\noindent\textbf{Sensitivity to {$\tau$} and $\lambda$.}
As shown in Fig.~\ref{fig:hyperparameters}, the model performance is unsatisfactory when the weight $\lambda$ is too high or too low. This might be because such values disrupt the balance of the overall loss. As expected, higher confidence of $\tau$ is preferable, aiding in filtering out unreliable pseudo-labels. The optimal performance is observed with $\lambda = 1$ and $\tau = 0.95$, adopted as the default setting for all experiments.
\begin{table}[t]
\centering
\captionsetup{font=footnotesize, justification=justified}

{\footnotesize
\begin{tabular*}{\linewidth}{@{\extracolsep{\fill}}l c cccc@{\extracolsep{\fill}}}
\toprule  
Method & Ratio & 1\textdegree\ vein & 2\textdegree\ vein & 3\textdegree\ vein & mIoU\\
\midrule  
\multirow{6}{*}{FixMatch } & 1:1:10 & 72.71 & 52.20 & 36.69 & 53.87\\
                           & 1:1:20 & 73.50 & 52.89 & 37.50 & 54.63\\
                           & 1:2:10 & 75.32 & 54.77 & 38.70 & 56.26\\
                           & 1:2:20 & 76.21 & 55.80 & 37.94 & 56.65\\
                           & 1:4:10 & 77.47 & 56.40 & 38.61 & 57.49\\
                           & 1:4:20 & 77.08 & 57.30 & 39.54 &57.97\\
\midrule  

\multirow{6}{*}{U$^2$PL} & 1:1:10 & 69.48 &46.46 &32.74 & 49.56\\
                         & 1:1:20 & 71.95 &49.67 &36.11 & 52.58\\
                         & 1:2:10 & 73.48 &51.02 &36.49 & 53.66\\
                         & 1:2:20 & 73.55 &51.30 &37.11 & 53.99\\
                         & 1:4:10 & 73.75 &52.50 &37.29 & 54.51\\
                         & 1:4:20 &74.15 &52.43 &37.92 & 54.83\\

\midrule  

\multirow{6}{*}{UniMatch} & 1:1:10 & 71.79 & 51.34 & 36.73 & 53.29\\
                          & 1:1:20 & 74.28 & 54.48 & 38.41 & 55.72\\
                          & 1:2:10 & 75.29 & 54.58 & 38.21 & 56.03\\
                          & 1:2:20 & 76.41 & 55.79 & 38.79 & 57.00\\
                          & 1:4:10 & 76.10 & 55.13 & 38.44  & 56.56\\
                          & 1:4:20 & 77.72 & 56.56 & 38.68 & 57.65\\

\bottomrule  
\end{tabular*}
}
\caption{Ablation study on different ratios of fully labeled, partially labeled, and unlabeled data under three SSL baselines with PSSS. Adding partially labeled data and/or unlabeled data can increase the performance. Each ratio unit contains 6 images with complete leaf blades.}
\label{tab:data_ratio}
\end{table}

\noindent\textbf{Sensitivity to Labeled Data Ratio.}
We study the impact of the relative ratio of partially labeled data and unlabeled data to fully labeled data.
As shown in Tab.~\ref{tab:data_ratio}, the model's performance can be improved when only raising the ratio of partially labeled or unlabeled data. Also, only a slight increase in the ratio of partially labeled data can result in a non-trivial performance gain. 

\begin{figure}[t]
    \centering
    \captionsetup{font=footnotesize, justification=justified}
    \begin{minipage}{\columnwidth}
        \centering
        \subcaptionbox{1\textdegree\ vein}{\includegraphics[width=0.49\linewidth]{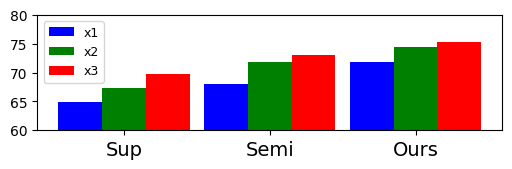}}
        \hfill
        \subcaptionbox{2\textdegree\ vein}{\includegraphics[width=0.49\linewidth]{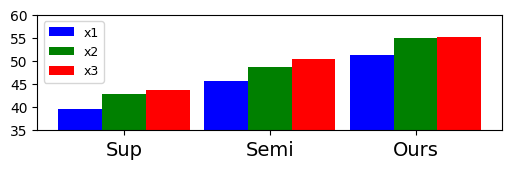}}
    \end{minipage}
    \begin{minipage}{\columnwidth}
        \centering
        \subcaptionbox{3\textdegree\ vein}{\includegraphics[width=0.49\linewidth]{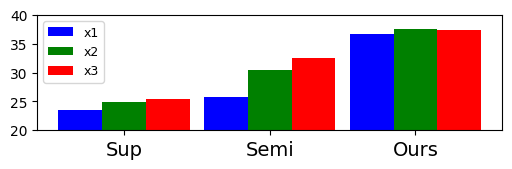}}
        \hfill
        \subcaptionbox{mIoU}{\includegraphics[width=0.49\linewidth]{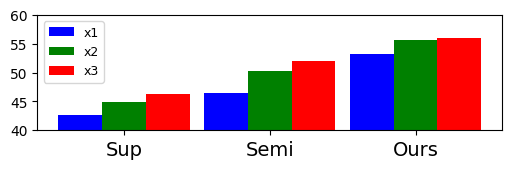}}
    \end{minipage}
    
    \caption{Comparison of performance across different data volumes. ``Sup'', ``Semi'',  and ``Ours'' refer to ``Supervised'', ``Unimatch'', and ``Unimatch + PSSS'', respectively. PSSS can efficiently utilize the labeled information, especially for the 3\textdegree\ vein.
    }
    \label{fig:data_scaling}

\end{figure}
\noindent\textbf{Effect of Dataset Size.}
We increase the volume of training data three times to study the scaling effect of dataset. 
As shown in Fig.~\ref{fig:data_scaling}, increasing data volume can achieve better performance under all three setups. However, there are two important findings. First, this scaling effect is not linear. There seems to be a diminishing return when the size of dataset becomes larger. 
Second, while increasing data volume results in some enhancement, the performance gain on segmenting the 3\textdegree\ vein remains limited. Importantly, though segmenting the 3\textdegree\ vein is difficult, this suggests that PSSS can use only $\frac{1}{3}$ of training data to reach the performance upper bound while significantly outperforming supervised and semi-supervised baselines.

\begin{table}[t]
\centering
\setlength{\tabcolsep}{4.5pt}
\captionsetup{font=footnotesize, justification=justified}
{\footnotesize
\begin{tabular}{@{\extracolsep{\fill}}lcccc@{\extracolsep{\fill}}}
\toprule
Backbone & 1\textdegree\ vein & 2\textdegree\ vein & 3\textdegree\ vein & mIoU\\
\midrule
ResNet101 & 68.07 & 45.57  & 22.32 & 45.32\\ 
ResNet101 + PSSS  & \textbf{72.71} & \textbf{52.20}  & \textbf{36.69} & \textbf{53.87}\\ 
\midrule
Xception & 69.71 & 45.83  & 19.51 & 45.02\\
Xception  + PSSS& \textbf{71.08} & \textbf{50.73}  & \textbf{34.49} & \textbf{52.10}\\
\midrule
Swin Transformer & 36.88 & 27.22  & 22.73 & 28.94\\
Swin Transformer + PSSS& \textbf{40.10} & \textbf{33.54}  & \textbf{25.04} & \textbf{32.89}\\
\bottomrule
\end{tabular}
}
\caption{Evaluation of PSSS under different network backbones. In this experiment, we use FixMatch as the SSL baseline. PSSS is robust under different network architectures.}
\label{table:backbone_ablation}
\end{table}

\noindent\textbf{Effect of Network Backbones.}
We employ DeepLabV3+ as the architecture with ResNet101 and Xception~\cite{chollet2017xception} serving as the backbones. Additionally, we select the Swin-Unet~\cite{cao2022swin} architecture that employs Swin Transformer~\cite{liu2021swin} as its backbone. 
As shown in Tab.~\ref{table:backbone_ablation}, the PSSS module can improve the model's performance on all three representative backbones, proving its superior generalization capabilities.

\begin{table}[th]
\centering
\captionsetup{font=footnotesize, width=\columnwidth, justification=justified}
{\footnotesize
\resizebox{0.47\textwidth}{!}{
\begin{tabular}{@{\extracolsep{\fill}}lcccccccc@{\extracolsep{\fill}}}
\toprule
\multirow{2}{*}{Method} & \multicolumn{4}{c}{Sweet cherry} & \multicolumn{4}{c}{London planetree} \\
\cmidrule{2-9} 
& 1\textdegree\ vein & 2\textdegree\ vein & 3\textdegree\ vein & mIoU & 1\textdegree\ vein & 2\textdegree\ vein & 3\textdegree\ vein & mIoU\\
\midrule
Sup & 29.30 & 23.51 & \phantom{0}9.62 & 20.81 & 26.43 & 12.46 & \phantom{0}7.83 & 15.57 \\
Semi & 35.31 & {\textbf{24.40}} & \phantom{0}8.14 & 22.62 & {\textbf{31.59}}& 13.71& \phantom{0}8.87& 18.06\\
Ours &{\textbf{59.40}}& 19.20 & {\textbf{13.16}} & {\textbf{30.59}} & 30.34 & {\textbf{26.07}} &{\textbf{26.94}} & {\textbf{27.78}}\\
\bottomrule
\end{tabular}%
}
}
\caption{Transfer learning performance from soybean to other two species. The baseline SSL method is UniMatch.}
\label{tab:cross_transfer}
\end{table}

\begin{table}[t!]
\centering
\captionsetup{font=footnotesize, width=\columnwidth, justification=justified}
{\footnotesize
\resizebox{0.47\textwidth}{!}{
\begin{tabular}{@{\extracolsep{\fill}}lcccccccc@{\extracolsep{\fill}}}
\toprule
\multirow{2}{*}{Method} & \multicolumn{4}{c}{Sweet cherry} & \multicolumn{4}{c}{London planetree} \\
\cmidrule{2-9} 
& 1\textdegree\ vein & 2\textdegree\ vein & 3\textdegree\ vein & mIoU & 1\textdegree\ vein & 2\textdegree\ vein & 3\textdegree\ vein & mIoU\\
\midrule
Semi & 25.24 & \phantom{0}3.43 &\phantom{0}0.00 &\phantom{0}9.56 & 35.45 & \phantom{0}1.32&\phantom{0}0.00 &12.25 \\
Ours & \textbf{35.02 } & \textbf{33.87} &\textbf{19.61} &\textbf{29.50} &\textbf{48.96} & \textbf{26.37}&\textbf{\phantom{0}1.46} &\textbf{25.60}\\
\midrule
\midrule
Oracle & 68.36 & 46.82 &13.25 &42.81 &67.67 &27.69 &\phantom{0}9.86 &35.07 \\
\bottomrule
\end{tabular}%
}
}
\caption{Cross-species learning performance from soybean to other two species. The baseline SSL method is UniMatch.}
\label{tab:cross_scarse}
\end{table}

\subsection{Analysis of Cross-species Learning}
\label{sec:cross-specie}
We conduct the first study of cross-species learning for leaf vein segmentation. Based on Tab.~\ref{table:dataset_annotation_time}, we find that soybean leaves are relatively labor-sparse and easy to be annotated in contrast to the other two species. Thus, soybean is considered the \emph{source} species, and the other two species are considered the \emph{target} species. In the first scenario, we assume that only soybean data are available and we perform direct transfer learning to two other species unseen in the training. We use the default ratio $1{:}1{:}10$ where the basic unit contains 20 images with full blades. The results are shown in Tab.~\ref{tab:cross_transfer}. Without seeing any samples from the target species, simply adding more partially labeled and unlabeled data from the source species can improve the overall model generalization on the target species, especially on the 3\textdegree\ vein.

In the second scenario, we consider an extreme label scarcity case: in addition to small-scale fully labeled source data, the target species can have access to unlabeled data and small-scale partially labeled data in the training. With the default ratio, there are 6 fully labeled source leaves, 6 partially labeled, and 60 unlabeled target leaves. The results are shown in Tab.~\ref{tab:cross_scarse}. We also include an ``Oracle'' for supervised training on fully labeled target species. We notice that though our method significantly outperforms the semi-supervised counterpart, the supervised Oracle can easily achieve better overall performance. We also conclude that, due to \emph{dataset shift}~\cite{quinonero2008covariate} between species, human annotations are still the key to successful applications. This further supports that HALVS is important for future research on the task of interest.

\section{Conclusion}
In this work, we explore the novel task of leaf hierarchical venation segmentation. We provide a finely annotated HALVS dataset for the first time and propose a label-efficient learning paradigm by considering the practical difficulties in annotating leaf veins. The empirical studies not only reveal new observations and challenges but also pose an insight into future research directions.

\clearpage
\section*{Acknowledgements}
This study was partially supported by the National Science Foundation of China (No.32200331, No.32090061), the Major Science and Technology Research Project of Hubei Province (No.2021AFB002), the Start-Up Grant from Wuhan University of Technology of China (No.104-40120526), and Shanghai Artificial Intelligence Laboratory.

\bibliographystyle{named}
\bibliography{IJCAI2024/refs}

\newpage
\appendix
\counterwithin{theorem}{section}
\counterwithin{figure}{section}
\numberwithin{equation}{section}

\section{Source Code}
The source code is provided in \texttt{\url{https://github.com/WeizhenLiuBioinform/HALVS-Hierarchical-Vein-Segment}} to re-implement the reported results, including detailed documentation of the data processing and training protocol.

\section{Additional Dataset Preparation Details}

\subsection{Data Collection}
The raw leaves of HALVS are collected by ten botany experts: soybean from the \textit{Soybean Experimental Station of Heilongjiang Academy of Agricultural Sciences, Mudanjiang, Heilongjiang Province, China}, sweet cherry from the \textit{Research Station of Shandong Institute of Pomology, Tai'an, Shandong Province, China}, and London planetree from the \textit{Wuhan University of Technology, Wuhan, Hubei Province, China}.

\subsection{Data Annotation}
\begin{figure}[ht]
    \centering
    \begin{subfigure}[b]{0.45\columnwidth}
        \includegraphics[width=\linewidth]{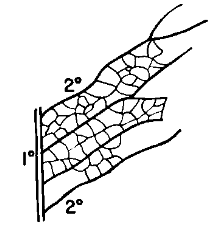}
        \caption{}
        \label{ann1}
    \end{subfigure}
     \hfill 
    \begin{subfigure}[b]{0.48\columnwidth}
        \includegraphics[width=\linewidth]{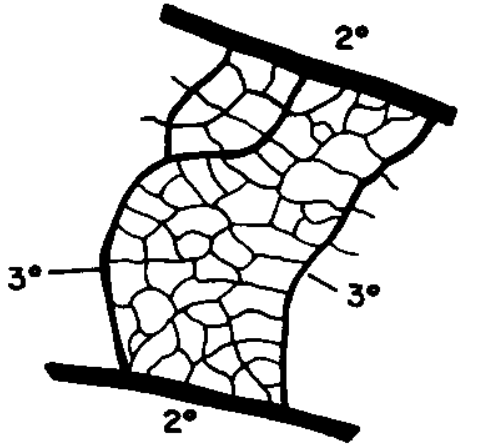}
        \caption{}
        \label{ann2}
    \end{subfigure}
    \caption{Illustration of three orders of veins (1\textdegree, 2\textdegree, and 3\textdegree). (a) and (b) are sourced from ~\protect\cite{hickey1973classification}.}
    \label{fig:images}
\end{figure}

We follow the academic rules of venation annotation ~\cite{hickey1973classification,ellis2009manual}. As shown in Fig.~\ref{ann1}, 1\textdegree\ veins are similar to tree trunks. They are usually the widest veins, extending from the base of the tree to the edges of the leaf. 2\textdegree\ veins typically extend outwards from the 1\textdegree\ veins, and their width is second only to that of the 1\textdegree\ veins. As shown in Fig.~\ref{ann2}, 3\textdegree\ veins are much narrower than the 2\textdegree\ veins, usually located between 1\textdegree\ and 2\textdegree\ veins, or between two 2\textdegree\ veins. It is worth noting that if they arise dichotomously or appear to have the same, or nearly the same, gauge as their parent vein, they are considered the same order as the source vein. 

\section{Additional Implementation Details}
To retain image details in our high-resolution data, we partition the original images into patches of size $256\times256$. Then, we eliminate patches that lack pixel information. The remaining patches are used for training. In the inference phase, we also use patches for prediction. The patches are assembled to acquire the final prediction of the original image size. 

\section{Additional Qualitative Analysis}
\begin{figure*}[htbp]
    \centering
    
    \begin{subfigure}{0.19\textwidth}
        \includegraphics[width=\linewidth]{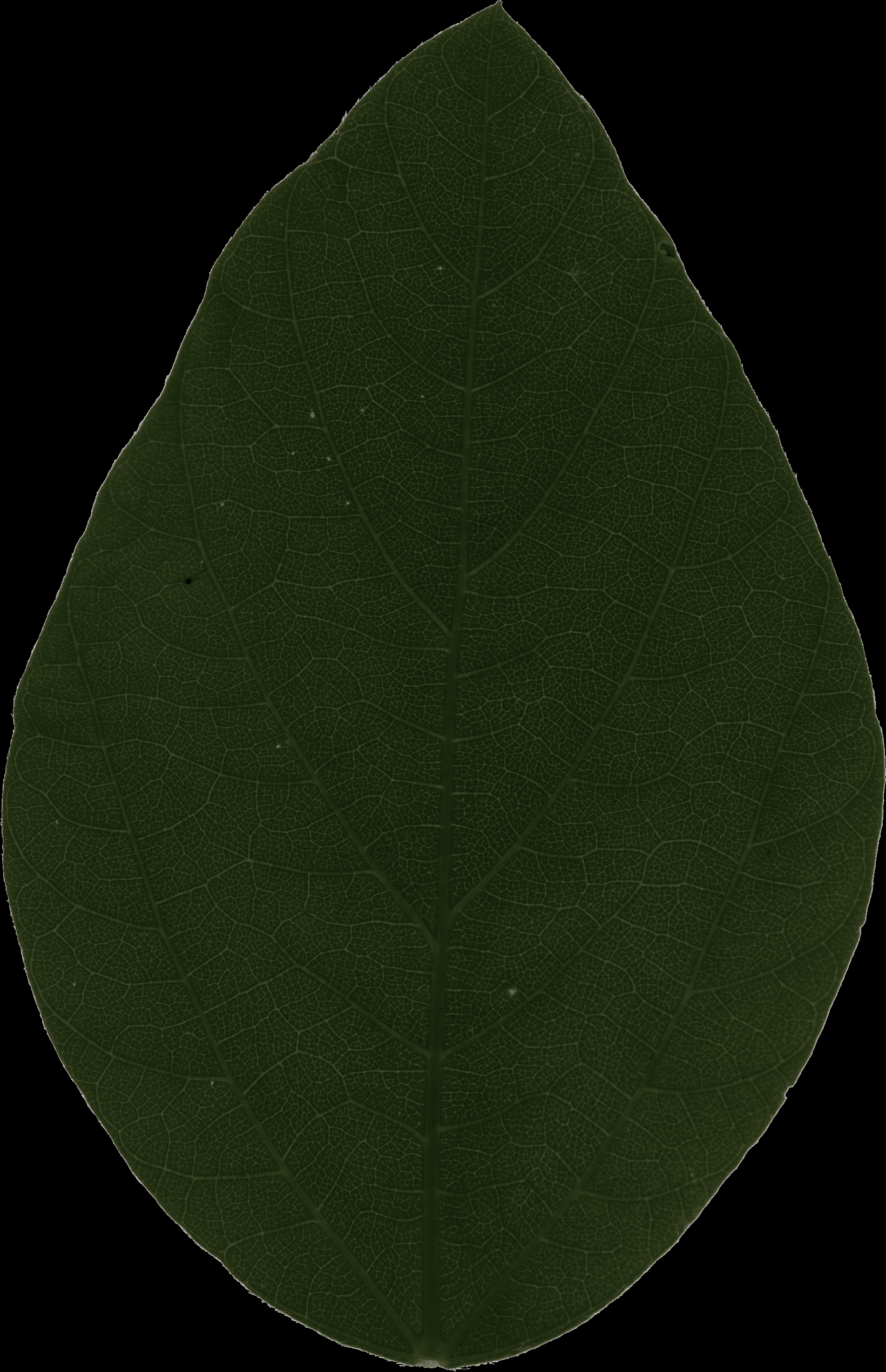} 
        \caption{Original Image}
        \label{fig:image2}
    \end{subfigure}
    \hfill
    \begin{subfigure}{0.19\textwidth}
        \includegraphics[width=\linewidth]{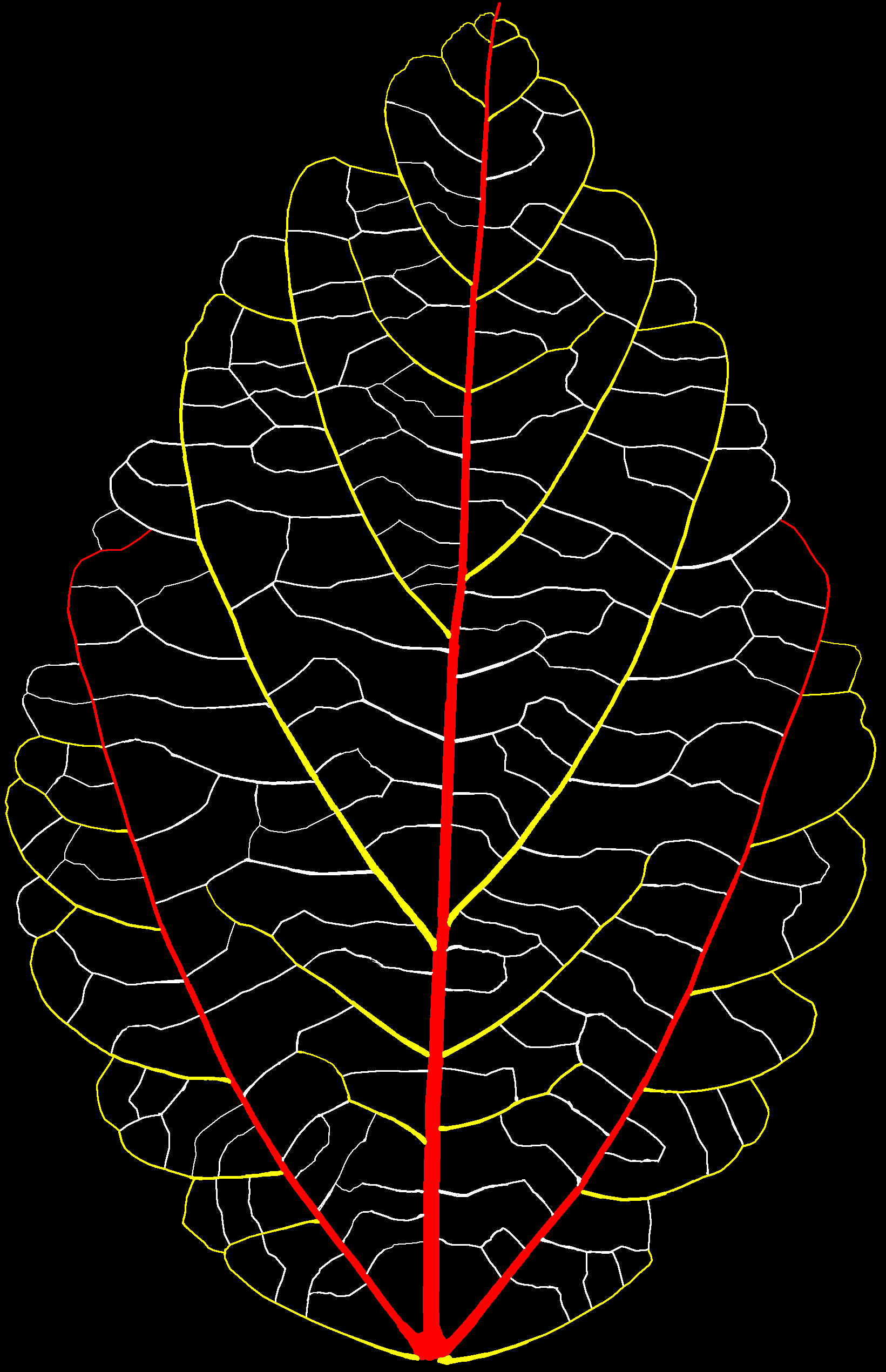}
        \caption{Ground Truth}
        \label{fig:gt}
    \end{subfigure}
    \hfill
    \begin{subfigure}{0.19\textwidth}
        \includegraphics[width=\linewidth]{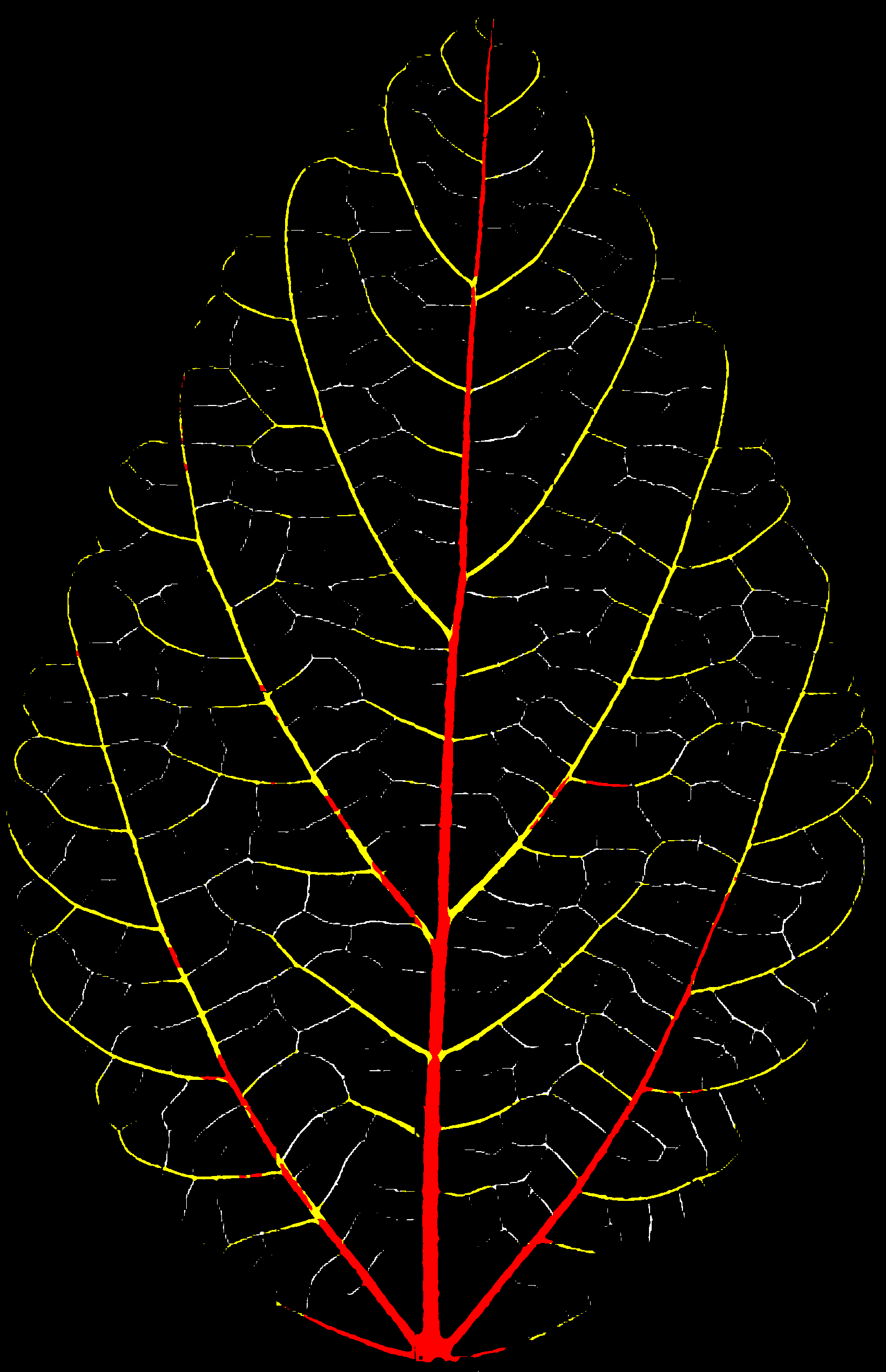}
        \caption{Supervised}
        \label{fig:sup}
    \end{subfigure}
    \hfill
    \begin{subfigure}{0.19\textwidth}
        \includegraphics[width=\linewidth]{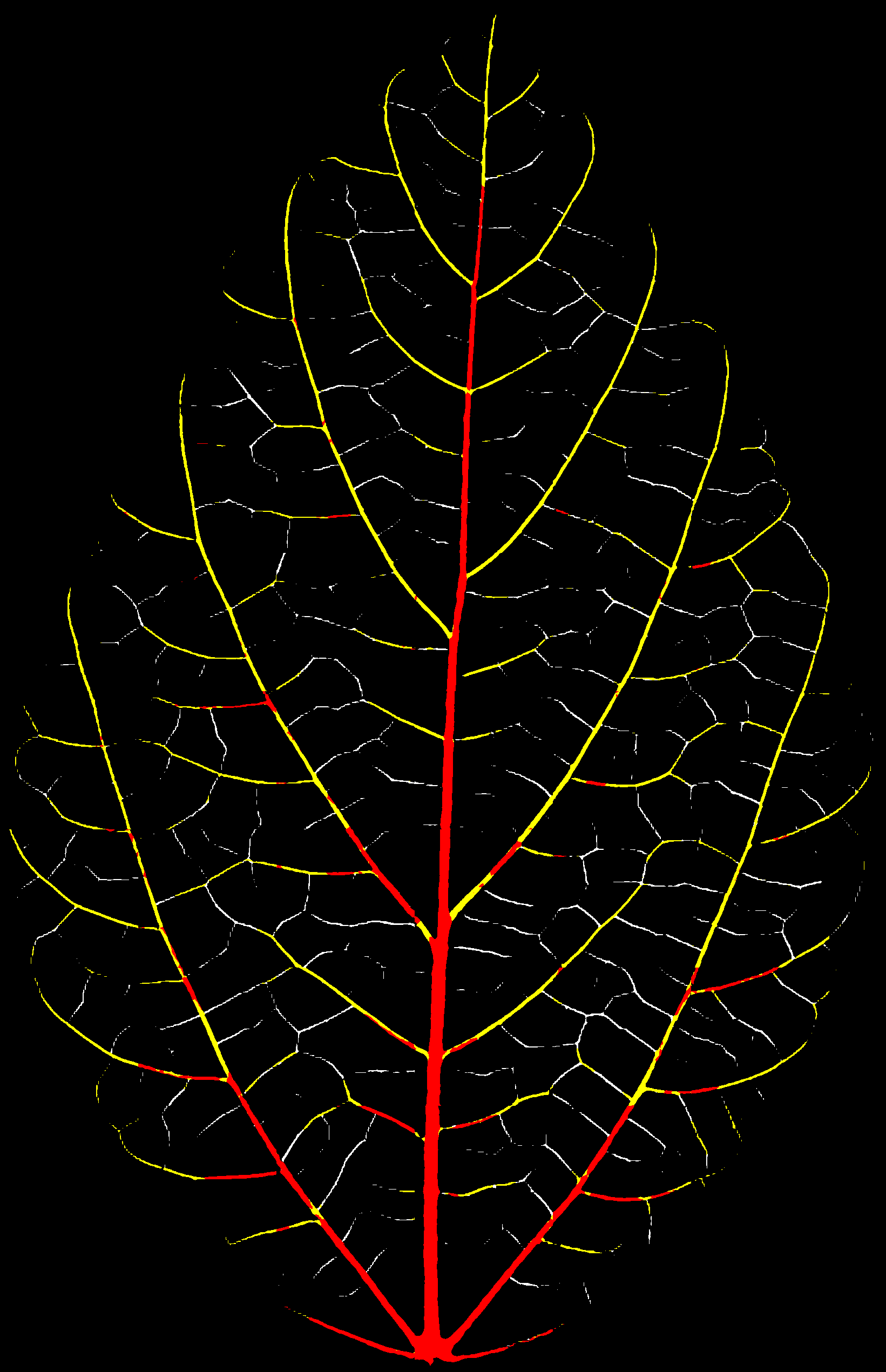}
        \caption{Semi-Supervised}
        \label{fig:semi}
    \end{subfigure}
    \hfill
    \begin{subfigure}{0.19\textwidth}
        \includegraphics[width=\linewidth]{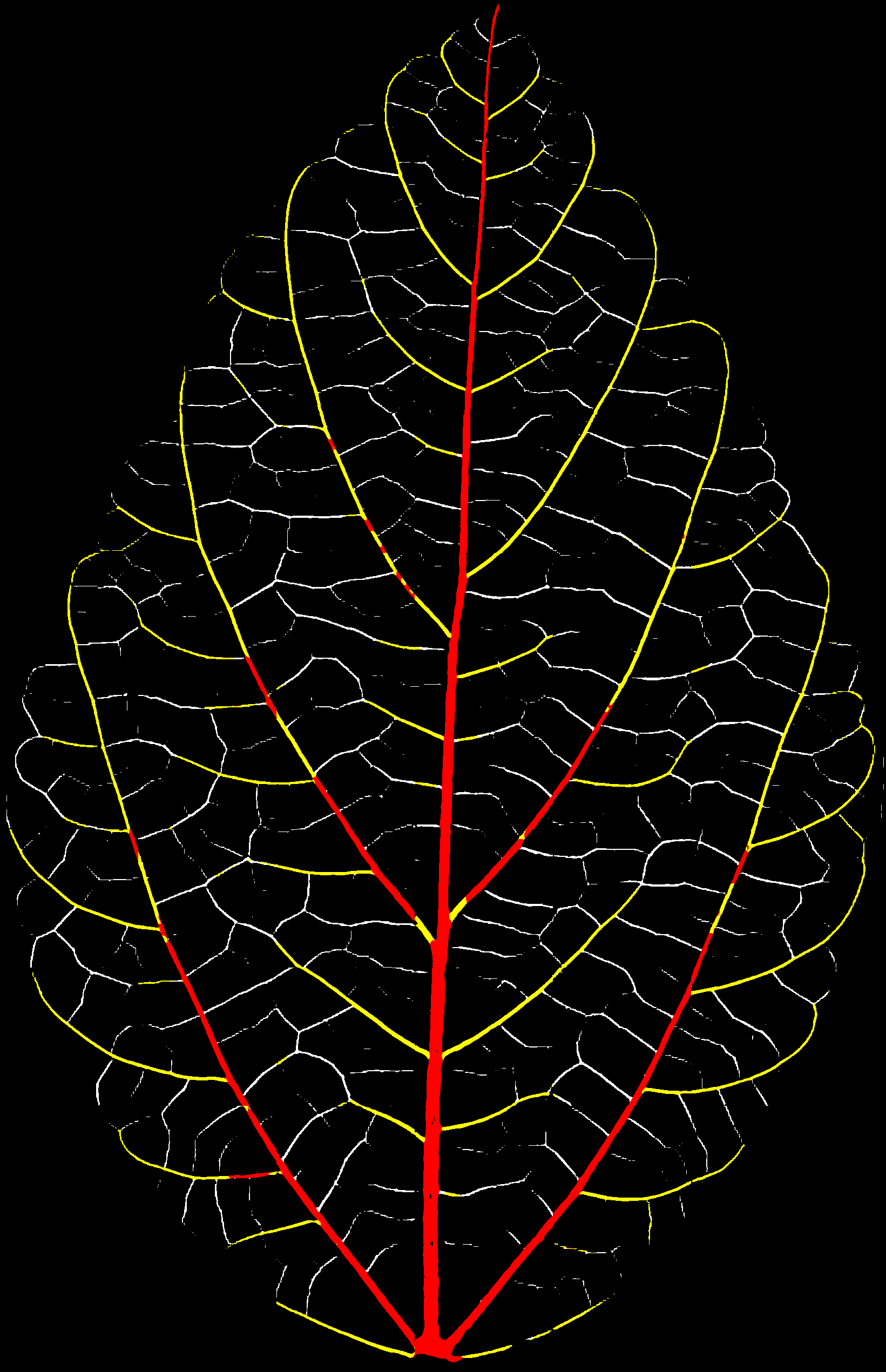}
        \caption{Ours}
        \label{fig:ours}
    \end{subfigure}
    
    \caption{Visualizations of the segmentation results on soybean leaf. The colors of red, yellow, and white represent 1\textdegree, 2\textdegree, and 3\textdegree\ veins, respectively. Under the same quantity of 3\textdegree\ vein annotations, the qualitative performance of the 3\textdegree\ vein in (e) is notably superior to its counterparts in (c) and (d). Best viewed with digital zoom.}
    \label{fig:soybean}
\end{figure*}

\begin{figure*}[htbp]
    \centering
    
    \begin{subfigure}{0.19\textwidth}
        \includegraphics[width=\linewidth]{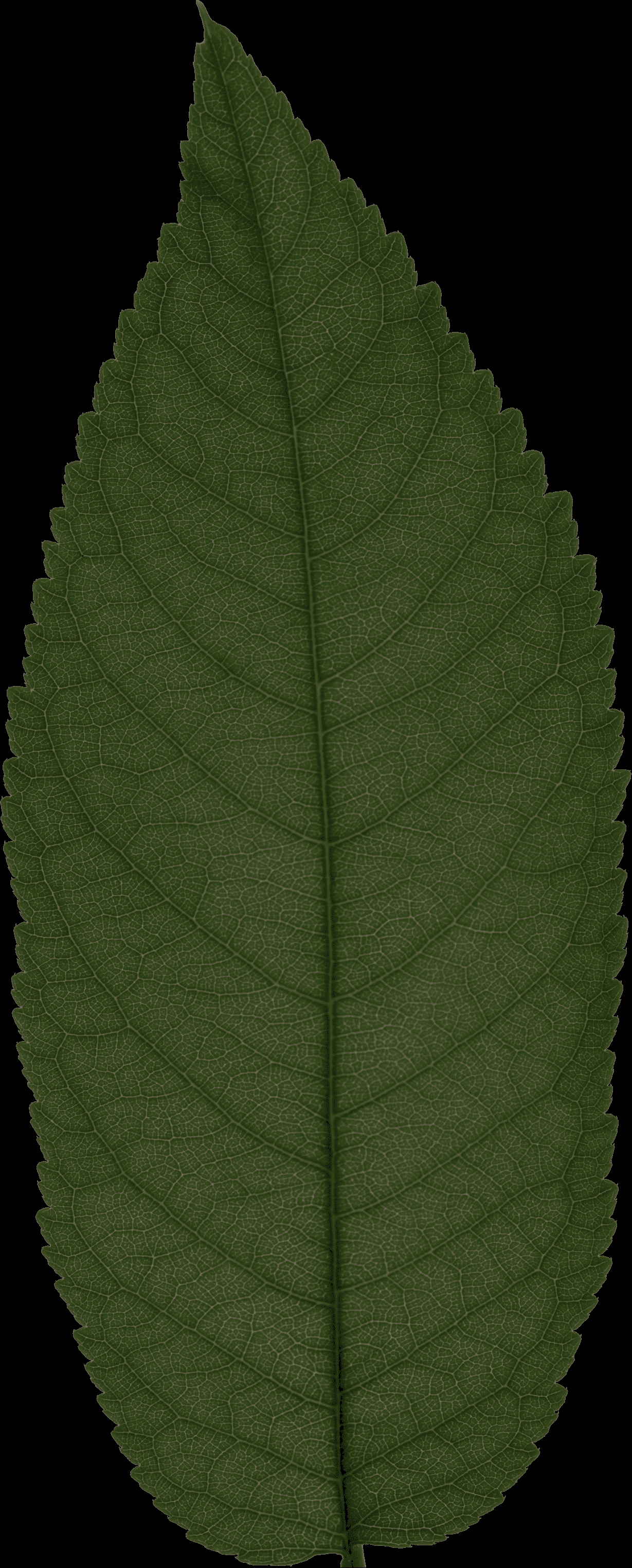} 
        \caption{Original Image}
        \label{fig:image3}
    \end{subfigure}
    \hfill
    \begin{subfigure}{0.19\textwidth}
        \includegraphics[width=\linewidth]{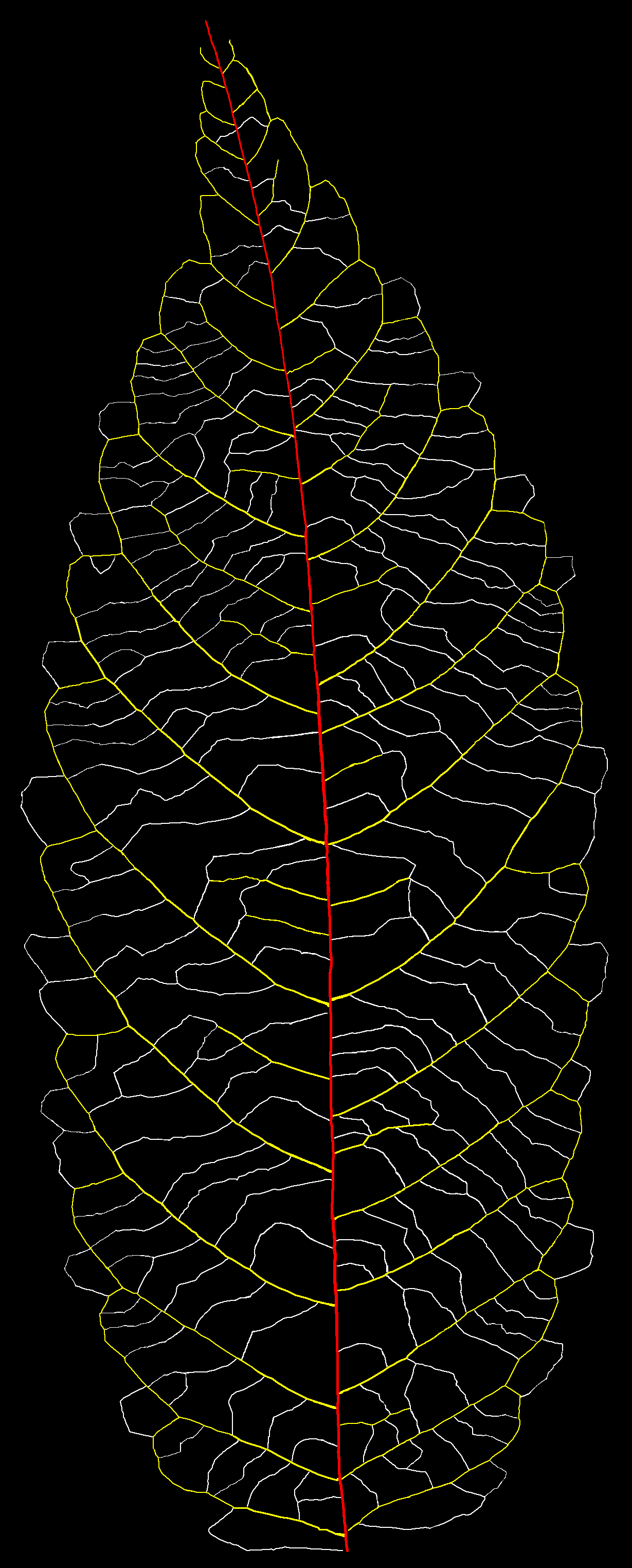}
        \caption{Ground Truth}
        \label{fig:gt}
    \end{subfigure}
    \hfill
    \begin{subfigure}{0.19\textwidth}
        \includegraphics[width=\linewidth]{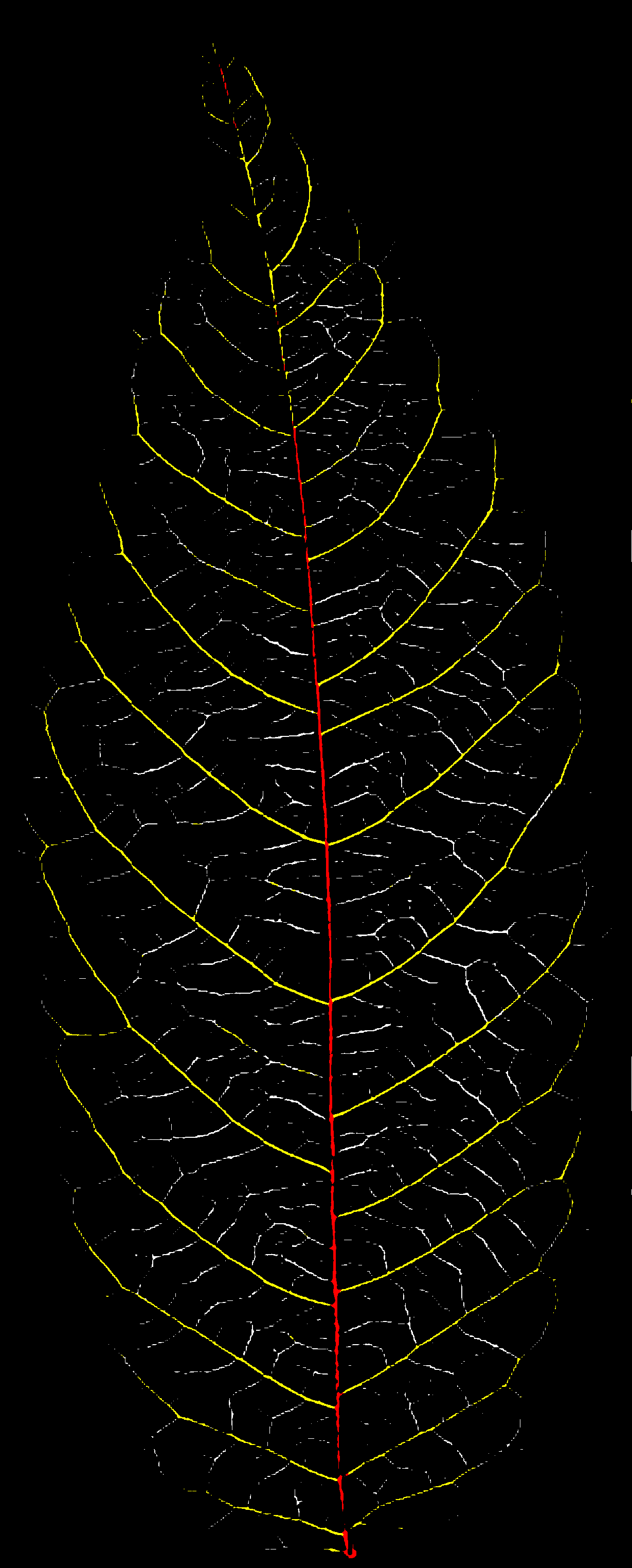}
        \caption{Supervised}
        \label{fig:sup}
    \end{subfigure}
    \hfill
    \begin{subfigure}{0.19\textwidth}
        \includegraphics[width=\linewidth]{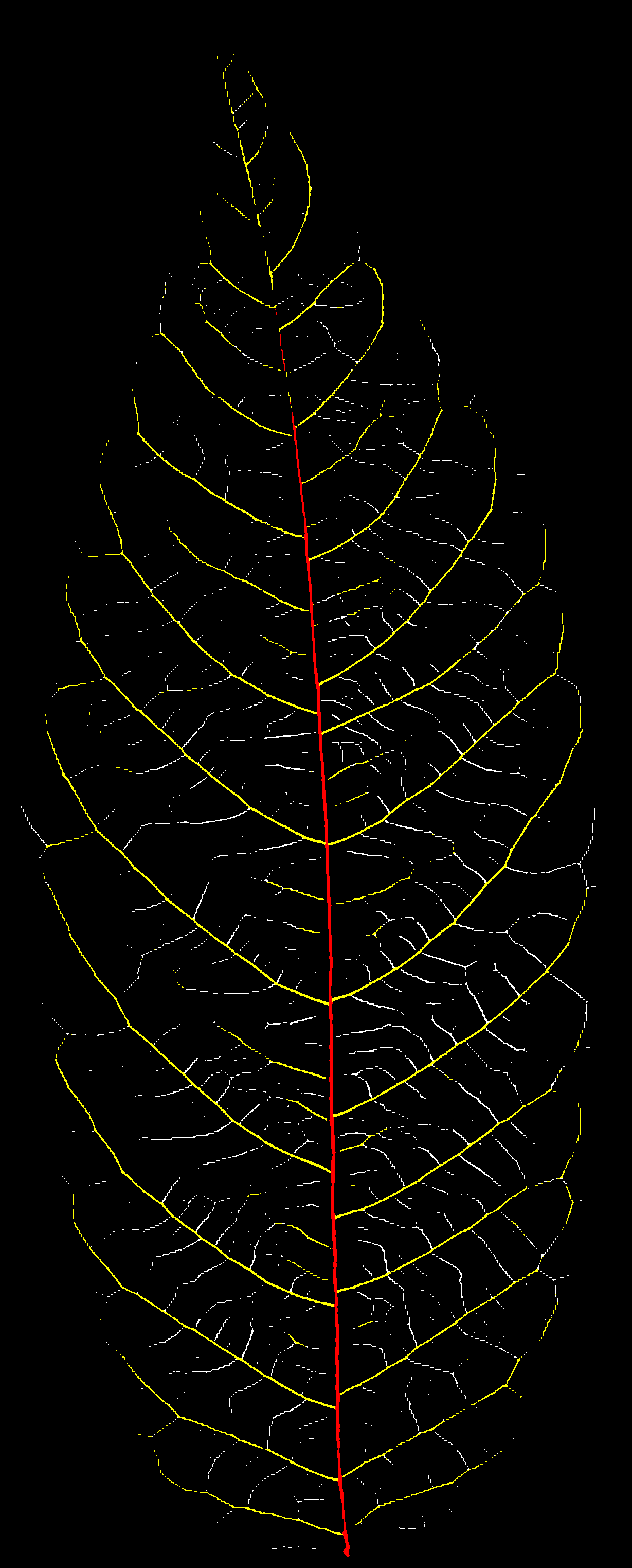}
        \caption{Semi-Supervised}
        \label{fig:semi}
    \end{subfigure}
    \hfill
    \begin{subfigure}{0.19\textwidth}
        \includegraphics[width=\linewidth]{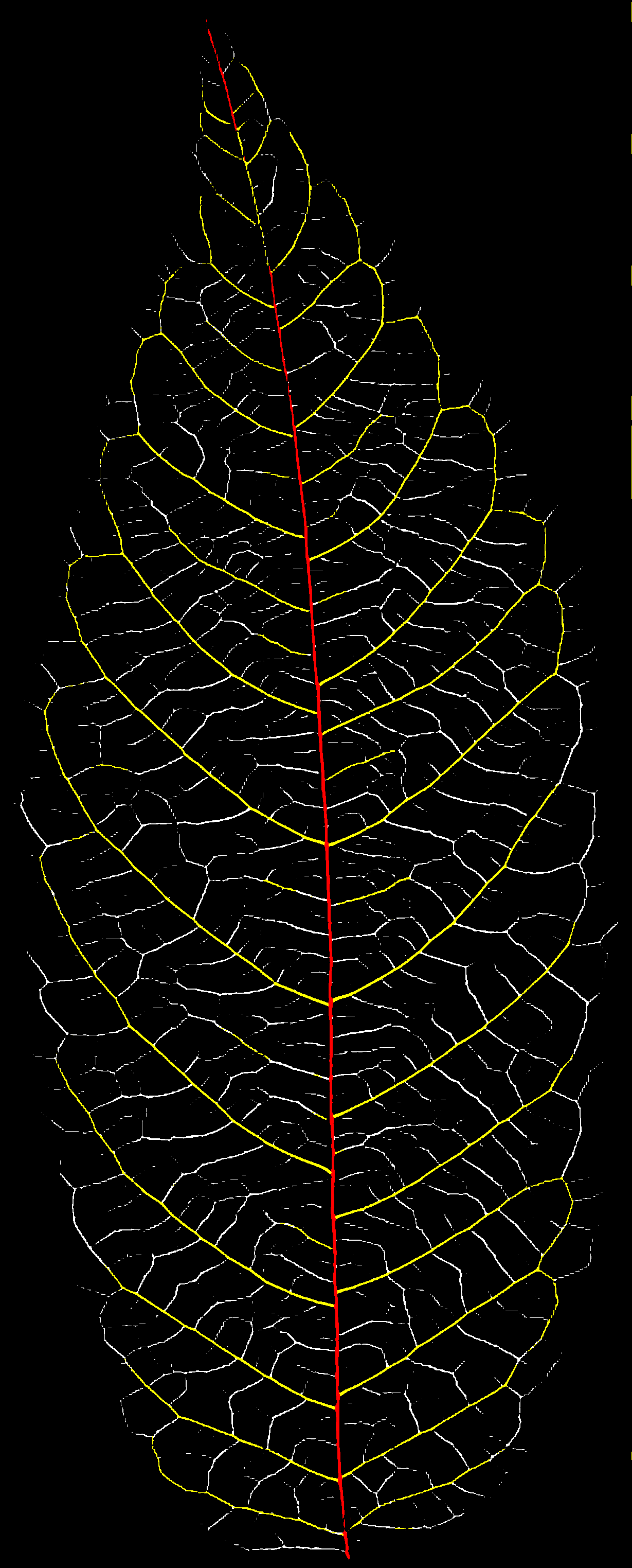}
        \caption{Ours}
        \label{fig:ours}
    \end{subfigure}
    
    \caption{Visualizations of the segmentation results on sweet cherry leaf. The colors of red, yellow, and white represent 1\textdegree, 2\textdegree, and 3\textdegree\ veins, respectively. Under the same quantity of 3\textdegree\ vein annotations, the qualitative performance of the 3\textdegree\ vein in (e) is notably superior to its counterparts in (c) and (d). Best viewed with digital zoom.}
    \label{fig:sweet_cherry}
\end{figure*}

\begin{figure*}[htbp]
    \centering
    \begin{subfigure}{0.35\textwidth}
        \includegraphics[width=\linewidth]{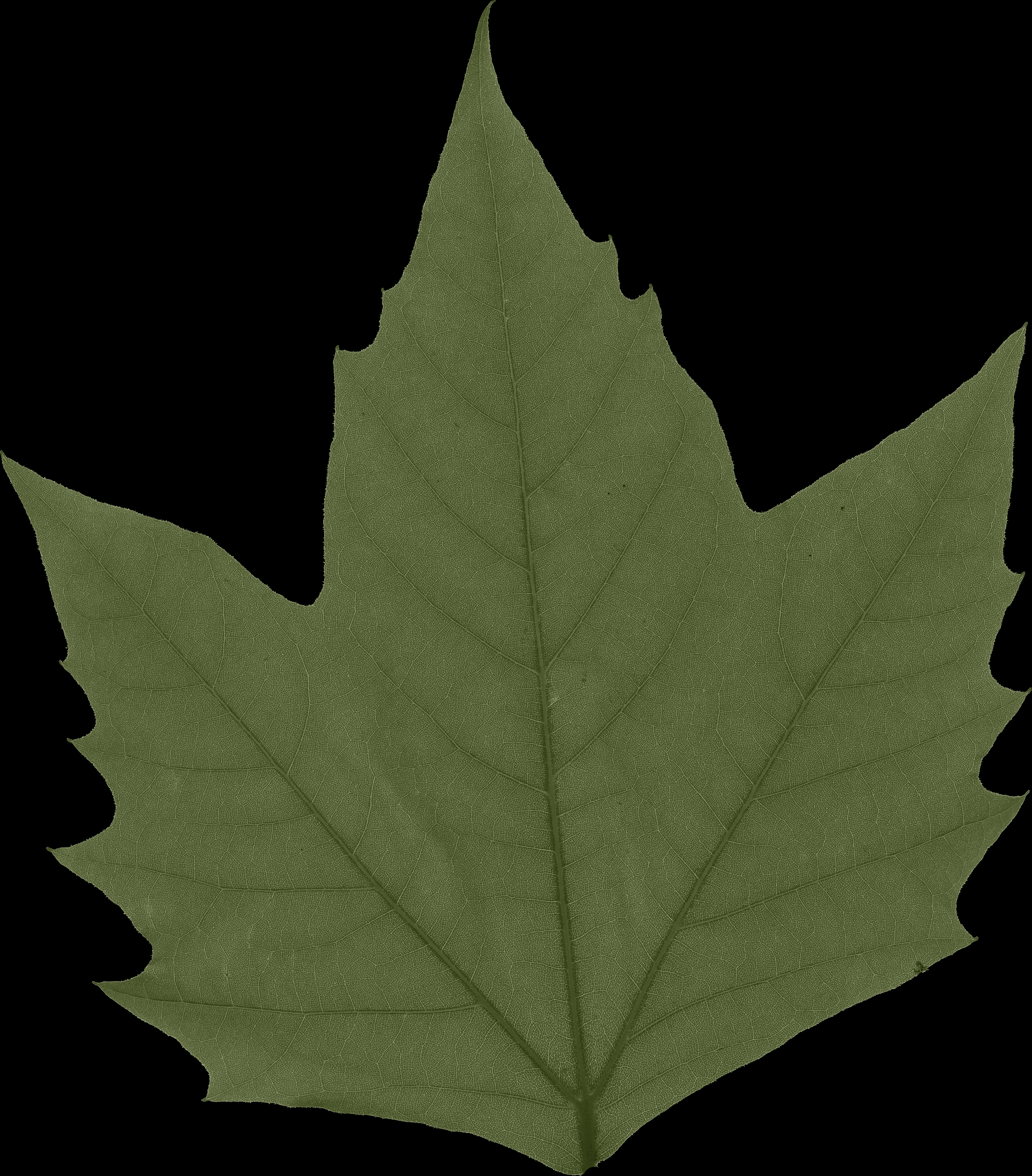}
        \caption{Original Image}
        \label{fig:sub1}
    \end{subfigure}
    
    \begin{subfigure}{0.35\textwidth}
        \includegraphics[width=\linewidth]{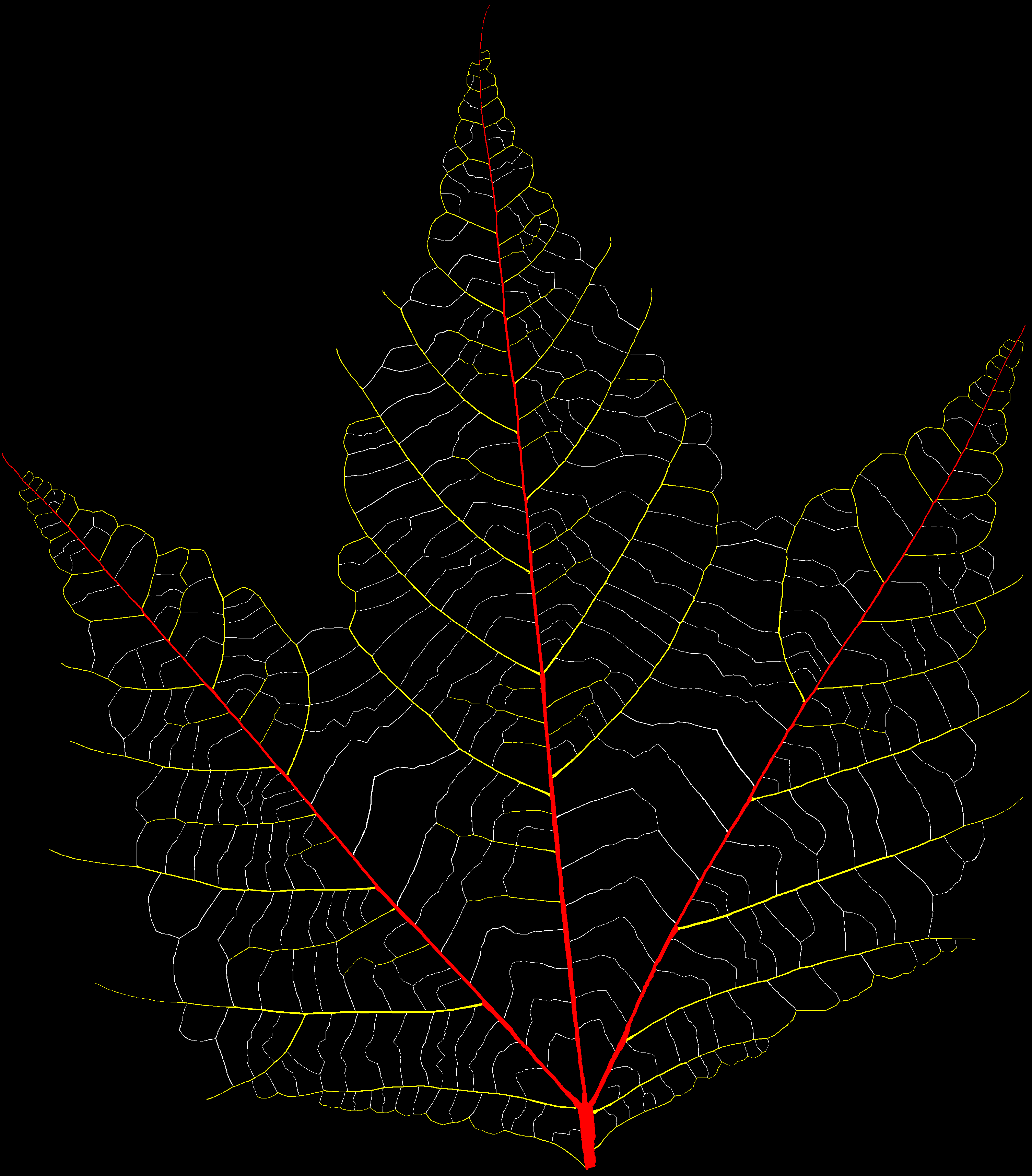}
        \caption{Ground Truth}
        \label{fig:sub2}
    \end{subfigure}
    \begin{subfigure}{0.35\textwidth}
        \includegraphics[width=\linewidth]{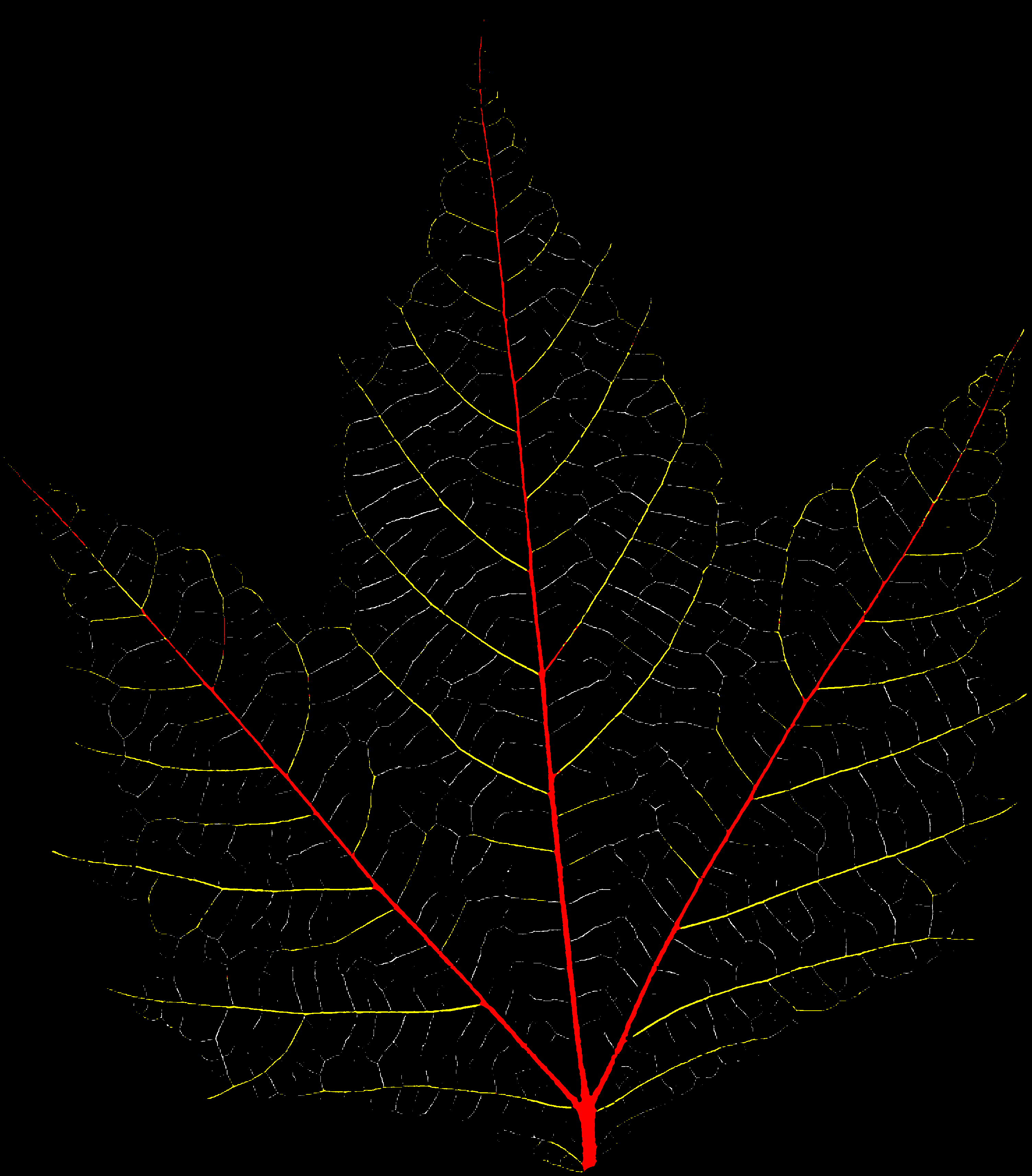}
        \caption{Supervised}
        \label{fig:sub3}
    \end{subfigure}
    
    \begin{subfigure}{0.35\textwidth}
        \includegraphics[width=\linewidth]{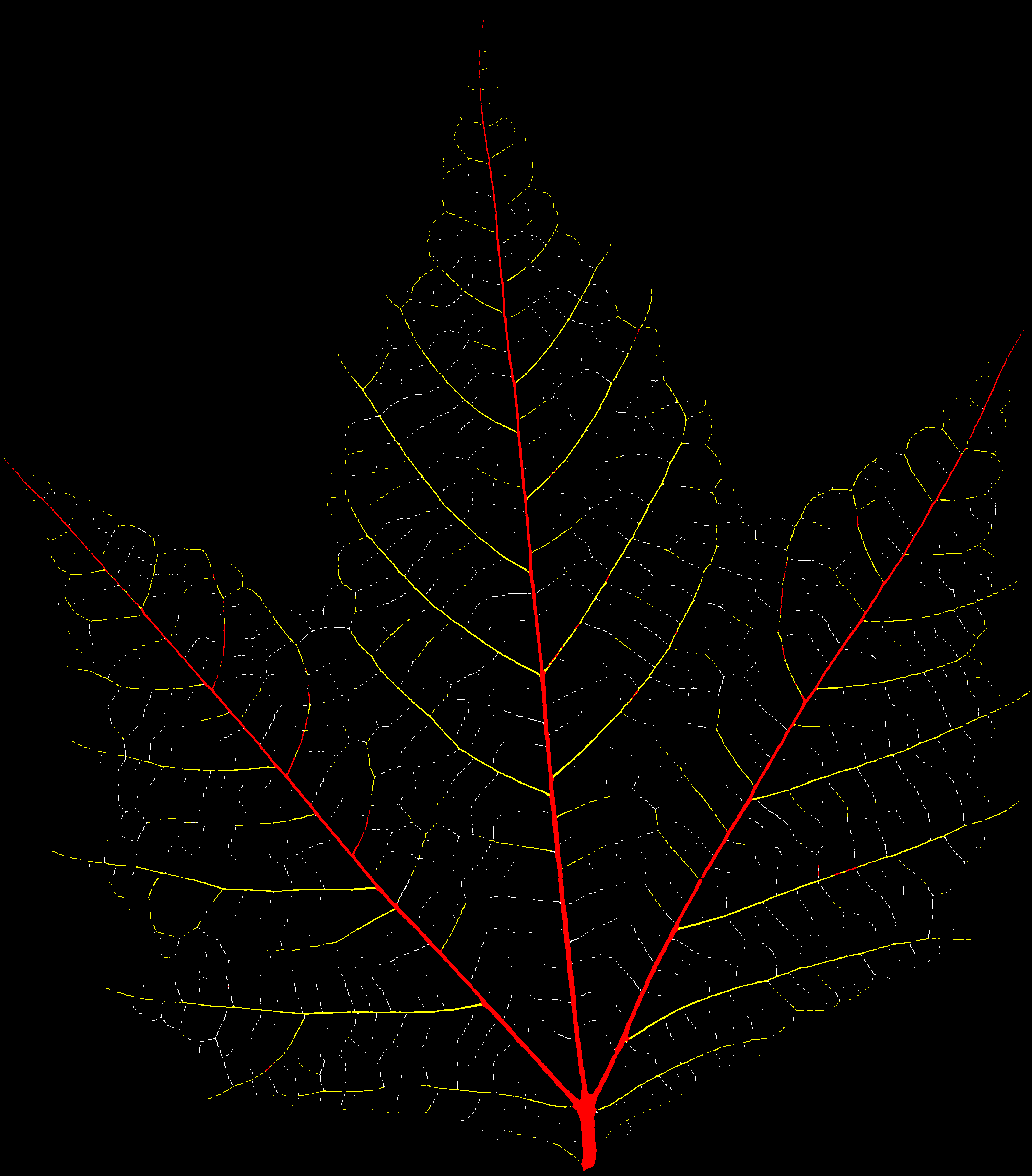}
        \caption{Semi-Supervised}
        \label{fig:sub4}
    \end{subfigure}
    \begin{subfigure}{0.35\textwidth}
        \includegraphics[width=\linewidth]{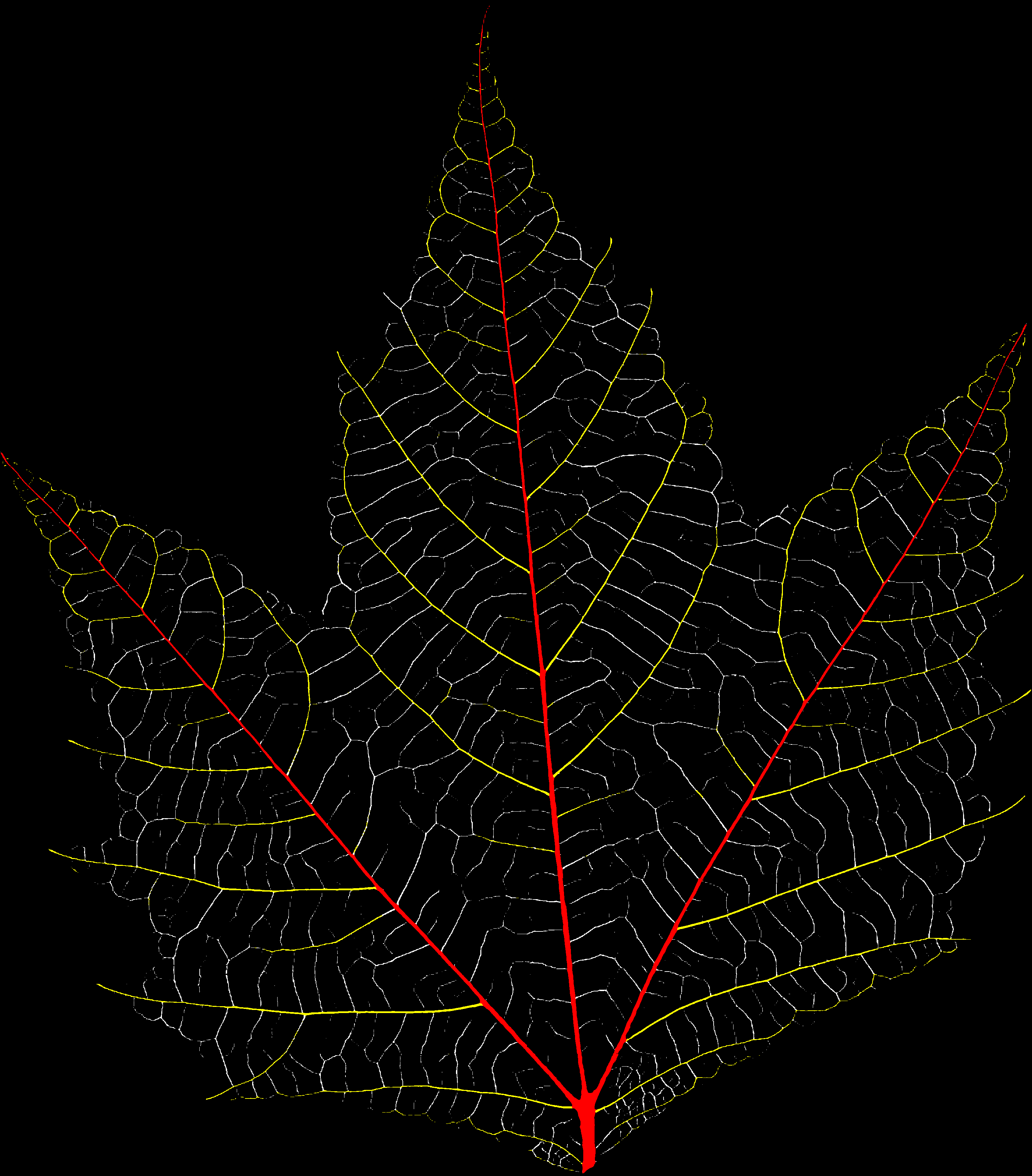} 
        \caption{Ours}
        \label{fig:new}
    \end{subfigure}

    \caption{Visualizations of the segmentation results on London planetree leaf. The colors of red, yellow, and white represent 1\textdegree, 2\textdegree, and 3\textdegree\ veins, respectively. Under the same quantity of 3\textdegree\ vein annotations, the qualitative performance of the 3\textdegree\ vein in (e) is notably superior to its counterparts in (c) and (d). Best viewed with digital zoom.}
    \label{planetree}
\end{figure*}

We provide additional qualitative comparisons between supervised baseline, semi-supervised baseline, and our partially supervised method, following the same setup described in Section 5.1 of the main text. As shown in Fig.~\ref{fig:soybean}, Fig.~\ref{fig:sweet_cherry}, and Fig.~\ref{planetree},
 our partially supervised method significantly outperforms the baseline methods when segmenting the 3\textdegree\ veins.

\end{document}